\documentclass[manuscript=article]{achemso}

\usepackage[version=3]{mhchem} 
\usepackage{graphicx}
\usepackage{hyperref}
\usepackage{upgreek}
\usepackage[T1]{fontenc}
\usepackage{textcomp}
\usepackage{amsmath}
\usepackage{caption}
\usepackage{subcaption}
\usepackage{algorithm}
\usepackage[noend]{algpseudocode}
\usepackage{float}
\usepackage{verbatim}
\usepackage{longtable}

\author{Matthew Ragoza}
\affiliation{Department of Neuroscience, University of Pittsburgh, Pittsburgh, PA}
\alsoaffiliation{Department of Computer Science, University of Pittsburgh, Pittsburgh, PA}
\author{Joshua Hochuli}
\affiliation{Department of Computer Science, University of Pittsburgh, Pittsburgh, PA}
\alsoaffiliation{Department of Biological Sciences, University of Pittsburgh, Pittsburgh, PA}
\author{Elisa Idrobo}
\affiliation{Department of Computer Science, The College of New Jersey, Ewing, NJ}
\author{Jocelyn Sunseri}
\author{David Ryan Koes}
\email{dkoes@pitt.edu}
\affiliation{Department of Computational and Systems Biology, University of Pittsburgh}

\title[CNN Scoring]
  {Protein-Ligand Scoring with Convolutional Neural Networks}


\begin{document}



\begin{abstract}
Computational approaches to drug discovery can reduce the time and cost associated with experimental assays and enable the screening of novel chemotypes. Structure-based drug design methods rely on scoring functions to rank and predict binding affinities and poses. The ever-expanding amount of protein-ligand binding and structural data enables the use of deep machine learning techniques for protein-ligand scoring.

We describe convolutional neural network (CNN) scoring functions that take as input a comprehensive 3D representation of a protein-ligand interaction. A CNN scoring function automatically learns the key features of protein-ligand interactions that correlate with binding. We train and optimize our CNN scoring functions to discriminate between correct and incorrect binding poses and known binders and non-binders. We find that our CNN scoring function outperforms the AutoDock Vina scoring function when ranking poses both for pose prediction and virtual screening.
\end{abstract}

\section{Introduction}
Protein-ligand scoring is a keystone of structure-based drug design.  Scoring functions rank and score protein-ligand structures with the intertwined goals of accurately predicting the binding affinity of the complex, selecting the correct binding mode (pose prediction), and distinguishing between binders and non-binders (virtual screening). 

Existing empirical \cite{sminapaper,Eldridge1997,Bohm1994score,Wang2002score,Korb2009,Friesner2004,Trott2009} and knowledge-based \cite{Huang2010,Muegge1999,Gohlke2000,Zhou2011, Mooij2005, Ballester2010} scoring functions parameterize a predetermined function, which is usually physically inspired, to fit data, such as binding affinity values.
Scoring functions that use machine learning \cite{Ashtawy2015,Sato2009,Ballester2010,zilian2013sfcscore,jorissen2005,schietgat2015predicting,sminapaper,deng2004,durrant2010nnscore,chupakhin2013predicting,durrant2011nnscore,durrant2015ml,gonczarek2016,wallach2015atomnet} 
provide greater flexibility and expressiveness as they learn both parameters and the model structure from data.
However, the resulting model often lacks interpretability, and the increased expressiveness increases the probability of overfitting the model to the data, in which case the scoring function will not generalize to protein targets or ligand chemotypes not in the training data.
The risk of overfitting increases the importance of rigorous validation \cite{Kramer2010,gabel2014beware}, but the inherent increase in flexibility allows machine learning methods to outperform more constrained methods when trained on an identical input set \cite{li2014importance}.  The choice of input features can limit the expressiveness of a machine learning method.  Features such as atom interaction counts \cite{durrant2011nnscore}, pairwise atom distance descriptors \cite{Ballester2010}, interaction fingerprints \cite{chupakhin2013predicting}, or ``neural fingerprints'' generated by learned atom convolutions \cite{gonczarek2016} necessarily eliminate or approximate the information inherent in a protein-ligand structure, such as precise spatial relationships.

Neural networks \cite{rojas2013neural} are a neurologically inspired supervised machine learning technique that is routinely and successfully applied to problems such as speech recognition and image recognition.  A basic network consists of an input layer, one or more hidden layers, and an output layer of interconnected nodes. Each hidden node computes a feature that is a function of the weighted input it receives from the nodes of the previous layer. The outputs are propagated to each successive layer until the output layer generates a classification.  The network architecture and choice of activation function for each layer determine the design of the network.  The weights that parameterize the model are typically optimized to fit a given training set of data to minimize the error of the network.

Deep learning \cite{lecun2015deep} refers to neural networks with many layers, which are capable of learning highly complex functions and have been made practical largely by the increase in computational power provided by modern graphics cards.
The expressiveness of a neural network model can be controlled by the network architecture, which defines the number and type of layers that process the input to ultimately yield a classification.  The network architecture can be manually or automatically tuned with respect to validation sets to be as expressive as needed to accurately model the data and reduce overfitting \cite{srivastava2014dropout,szegedy2015going}.
Structure-based scoring functions that use neural networks \cite{durrant2010nnscore,chupakhin2013predicting,durrant2011nnscore,durrant2015ml,gonczarek2016,wallach2015atomnet} were recently shown to be competitive with empirical scoring in retrospective virtual screening exercises while also being effective in a prospective screen of estrogen receptor ligands \cite{durrant2015estrogen}.  Neural networks have also been successfully applied in the cheminformatics domain through creative manipulations of 2D chemical structure and construction of the network architecture \cite{xu2015deep,Lusci2013,DuvMacetal15nfp,ramsundar:2015}.

\begin{figure}[tbp]
\includegraphics[width=\linewidth]{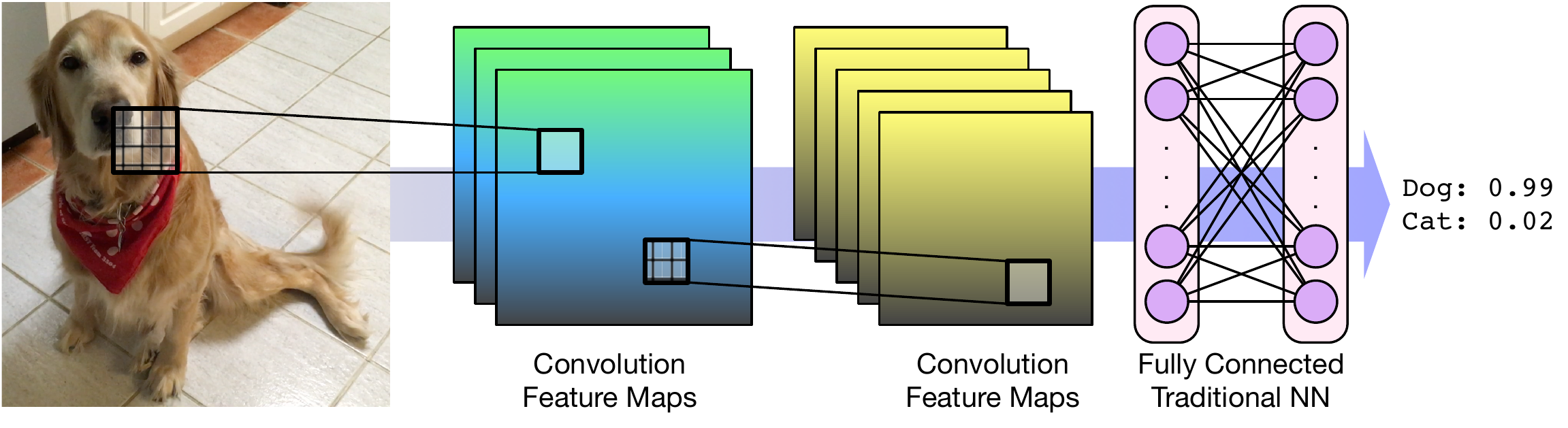}
\caption[]{\label{lily} A classical convolutional neural network for image recognition.  The first layer applies three different convolutions to the input image to create three maps of low level features that are the input for another convolutional layer that creates five maps. Feature maps preserve the spatial locality of the features. As a last step, a traditional neural net is applied to generate a classification.
}
\end{figure}

Convolutional neural networks (CNNs) \cite{lecun2015deep} are a type of neural network commonly used in image recognition.  CNNs hierarchically decompose an image so that each layer of the network learns to recognize higher-level features while maintaining their spatial relationships as illustrated in Figure~\ref{lily}.  For example, the first layer may learn to identify lines and corners in an image, the next may assemble these features to learn different shapes, and so on until the final layer can recognize something as high-level and complex as a dog breed.
CNNs are the best performing method for image recognition \cite{krizhevsky2012imagenet}, as epitomized by the GoogLeNet winning entry to the ImageNet Large Scale Visual Recognition Challenge of 2014 \cite{szegedy2015going} and the Microsoft ResNet entry of 2015 \cite{msdeepresidual}, both of which perform better at classifying images than most humans \cite{googlenetaccuracy}.

The impressive performance of CNNs at the image recognition task suggests that they are well-suited for learning from other types of spatial data, such as protein-ligand structures. Unlike previous machine learning methods, a CNN scoring method does not require the extraction of features from the structure. Instead, the method automatically identifies the most informative features required for successful scoring. This allows for the extraction of features that are not readily encoded in simplified potentials, such as hydrophobic enclosure \cite{Friesner2006} or surface area dependent terms \cite{Jain1996}, as well as features that have not yet been identified as relevant by any existing scoring function. 

Here we describe the development of a CNN model for protein-ligand scoring that is trained to classify compound poses as binders or non-binders using a 3D grid representation of protein-ligand structures generated through docking.  We show that our CNN scoring method outperforms the AutoDock Vina \cite{Trott2009} scoring function that is used to generate the poses both when selecting poses for pose prediction and for virtual screening tasks.  We also illustrate how our CNN score can be decomposed into individual atomic contributions to generate informative visualizations.

\section{Methods}

In order to create our CNN scoring models we utilize two training sets, one focused on pose prediction and the other on virtual screening.  The structural information in these sets is translated into a custom input format appropriate for CNN processing.  We systematically optimize the network topology and parameters using clustered cross-validation. The optimized network is then trained on the full training set and evaluated with respect to independent test sets. The predictions from the resulting models are decomposed into atomic contributions to provide informative visualizations.

\subsection{Training Sets}

We utilize two training sets focused on two different goals: pose prediction and virtual screening.  In all cases we generate ligand poses for actives and decoys using docking with smina\cite{sminapaper} and the AutoDock Vina scoring function \cite{Trott2009}.  We use docked poses, even for active compounds with a known crystal structure, because (1) these are the types of poses the model will ultimately have to score and (2) to avoid the model simply learning to distinguish between docked poses and crystal structures (which were likely optimized with different force fields).

Ligands are docked against a reference receptor within a box centered around a reference ligand with 8{\AA} of padding.  If 3D coordinates are not available for the ligand, a single 3D conformer of the ligand is generated using RDKit \cite{rdkit}  to provide the initial coordinates (using \texttt{rdconf.py} from \url{https://github.com/dkoes/rdkit-scripts}).  A single conformer is sufficient since the docking algorithm will sample the degrees of freedom of the ligand.  All docking is done against a rigid receptor that is stripped of water but not metal ions.  Protonation states for both the ligand and receptor are determined using OpenBabel \cite{openbabelpaper}. 

\subsubsection{Pose Prediction: CSAR}

Our pose prediction training set is based on the CSAR-NRC HiQ dataset, with the addition of the CSAR HiQ Update \cite{csarsel}.  This set consists of 466 ligand-bound co-crystals of distinct targets.  To generate the training set, we re-docked these ligands with the settings \texttt{--seed 0 --exhaustiveness 50 --num\_modes 20} to thoroughly and reproducibly sample up to 20 distinct poses.  We exclude targets where the ligand is annotated with a binding affinity of less than 5 pK units (a value provided as part of the CSAR dataset).  This results in 337 co-crystals where the ligand has a reported binding affinity better than 10$\mu$M (where the affinity may come from a variety of sources, including IC50 measurements).
For the purposes of training, poses with a heavy-atom RMSD less than 2{\AA} from the crystal pose were labeled as positive (correct pose) examples and those with an RMSD greater than 4{\AA} RMSD were labeled as negative examples.  Poses with RMSDs between 2{\AA} and 4{\AA} were omitted.  The final training set consists of 745 positive examples from 327 distinct targets and 3251 negative examples from 300 distinct targets (some targets produce only low or high RMSD poses).

\subsubsection{Virtual Screening: DUD-E}

Our virtual screening training set is based off the Database of Useful Decoys: Enhanced (DUD-E) \cite{Mysinger2012} dataset.
DUD-E consists of 102 targets, more than 20,000 active molecules, and over one million decoy molecules.  Unlike the CSAR set, crystal poses of these ligands are not provided, 
although a single reference complex is made available.  To generate poses for training, we dock against this reference receptor using smina's default arguments for exhaustiveness and sampling and select the pose that is top-ranked by the AutoDock Vina scoring function.  Top-ranked poses are used both for the active and decoy compounds. The result is an extremely noisy and unbalanced training set. The noisiness stems from cross-docking ligands into a non-cognate receptor, which substantially reduces the retrieval rate of low-RMSD poses in a highly target-dependent manner \cite{sminapaper}, as well as the use of randomly chosen decoys in DUD-E (the dataset may contain false negatives).  The unbalance is due to the much larger number of decoy molecules.  The final training set contains 22,645 positive examples and 1,407,145 negative examples.

\subsection{Input Format}

\begin{table}[tbp]
\begin{tabular}{c|c|c}
Type & Ligand & Receptor \\ \hline \hline
AliphaticCarbonXSHydrophobe & Y & Y \\
AliphaticCarbonXSNonHydrophobe & Y & Y \\
AromaticCarbonXSHydrophobe & Y & Y \\
AromaticCarbonXSNonHydrophobe & Y &  Y\\
Bromine & Y &  N \\
Calcium & N & Y \\
Chlorine & Y & N \\
Fluorine & Y &  N\\
Iodine & Y & N \\
Iron & N & Y \\
Magnesium & N & Y \\
Nitrogen & Y &  Y \\
NitrogenXSAcceptor & Y & Y \\
NitrogenXSDonor & Y & Y \\
NitrogenXSDonorAcceptor & Y & Y \\
Oxygen & Y & N \\
OxygenXSAcceptor & Y & Y \\
OxygenXSDonorAcceptor & Y & Y \\
Phosphorus & Y & Y \\
Sulfur & Y & Y \\
SulfurAcceptor & Y & N \\
Zinc & N & Y \\
\end{tabular}                
\caption{\label{atomtypes} Atom types used to define protein-ligand structures for CNN scoring. 
 }
\end{table}

\begin{figure}[tbp]
\includegraphics[width=0.5\linewidth]{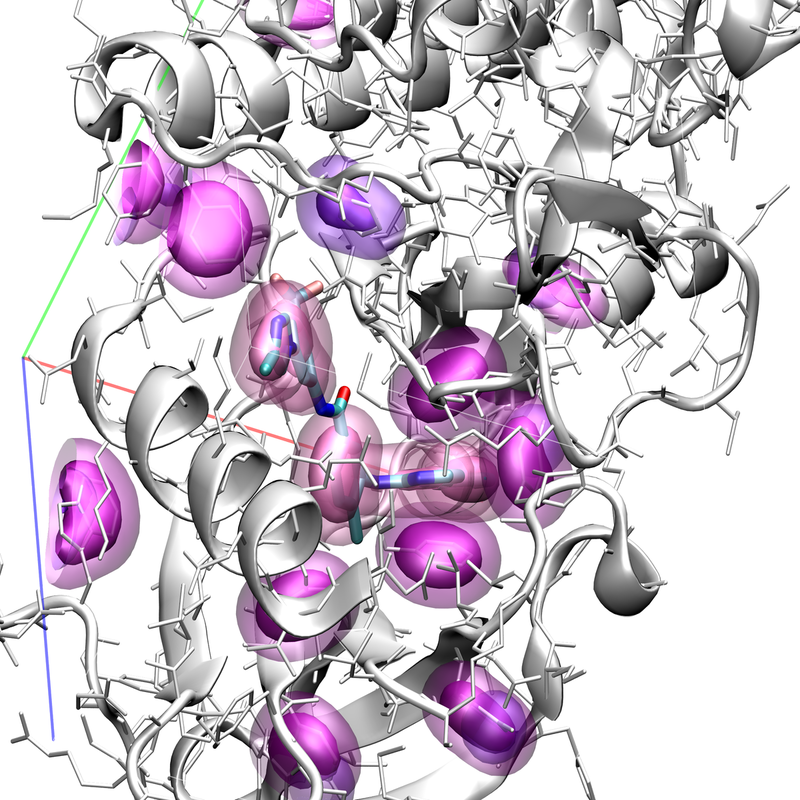}
\caption[]{\label{aromatic} Visualization of atom densities used as input to CNN scoring.  Aromatic carbon atom densities are shown at two isosurface levels (solid and transparent surfaces) for both the receptor (purple) and ligand (lavender).}
\end{figure}

Traditionally, CNNs take images as inputs, where a scene is discretized into pixels with red, green, and blue values (RGB). To handle our 3D structural data, we discretize a protein-ligand structure into a grid. The grid is 24{\AA}$^3$ and centered around the binding site with a default resolution of 0.5{\AA}, although we evaluate alternative resolutions.
Each grid point stores information about the types of heavy atoms at that point.  
Ligand and protein atoms have distinct atom types and each atom type is represented in a different channel (analogous to RGB channels in images) of the 3D grid.  Our default is to use smina \cite{sminapaper} atom types for a total of 34 distinct types with 16 receptor types and 18 ligand types as shown in Table~\ref{atomtypes}.
Only smina atom types that were present in the ligands and proteins of the training set were retained.  For example, halogens are not included as receptor atom types and metals are not included as ligand atom types.
Hydrogen atoms are ignored except to determine acceptor/donor atom types.  We also evaluate alternative atom typing schemes.
Atom type information is represented as a density distribution around the atom center.  We represent each atom as a function $A(d, r)$ where $d$ is the distance from the atom center and $r$ is the van der Waals radius:
\begin{equation}
\label{atom_gridder}
A(d, r) = 
\begin{cases}
e^{-\frac{2{d}^2}{{r}^2}} & 0 \leq d < r \\
\frac{4}{e^2r^2}{d}^2 - \frac{12}{e^2r}d + \frac{9}{e^2} & r \leq d < 1.5r \\
0 & d \geq 1.5r \\
\end{cases}
\end{equation}

$A$ is a continuous piecewise combination of a Gaussian (from the center to the van der Waals radius) and a quadratic (which goes to zero at 1.5 times the radius). This provides a continuous representation of the input. We also evaluate a `hard' discrete boolean representation.

 We generate these grids of atom density using a custom, GPU-accelerated layer, \texttt{MolGrid\-DataLayer}, of the Caffe\cite{jia2014caffe} deep learning framework.  This layer can process either standard molecular data files, which are read using OpenBabel \cite{openbabelpaper}, or a compact, custom binary \texttt{gninatypes} file that contains only the atomic coordinates and pre-processed atom type information. 

A visualization of our atom type volumetric representation is shown in Figure~\ref{aromatic} with density data rendering using isosurfaces.  This input format fully represents the spatial and chemical features of the protein-ligand complex; the sole approximations are the choice of grid resolution and the atom typing scheme. 

\subsection{Training}
\begin{figure}[tbp]
\includegraphics[width=0.6\linewidth]{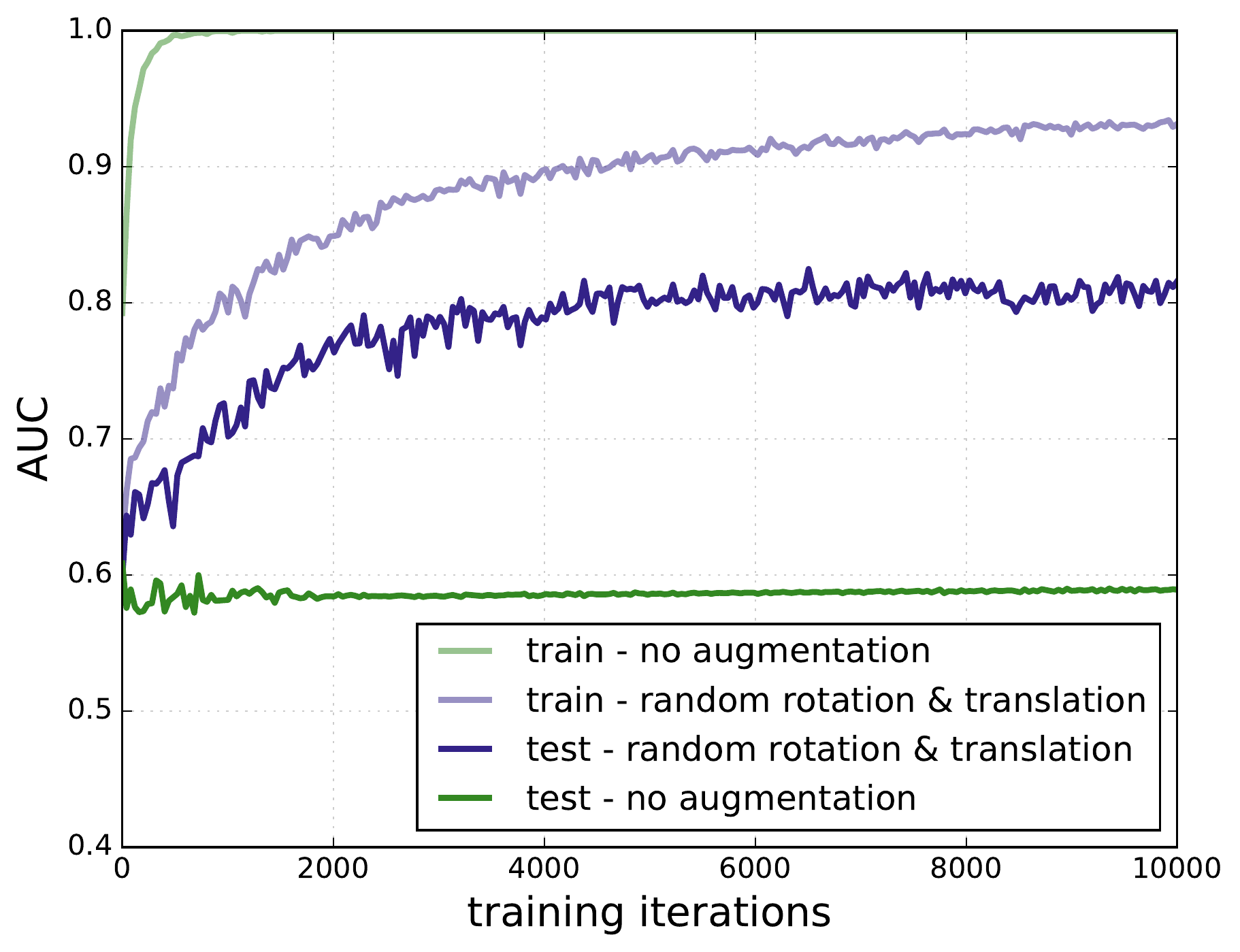}
\caption[]{\label{depth3_train} AUC on training and test sets, with and without data augmentation. Training on CSAR without data augmentation results in classic signs of overfitting: the training set AUC approaches 1.0, but the test AUC plateaus at a much lower value. When additional random rotations and translations are included in the training set, overfitting is reduced.}
\end{figure}

Our CNN models were defined and trained using the Caffe deep learning framework \cite{jia2014caffe}.  Training minimized the multinomial logistic loss of the network using a variant of stochastic gradient descent (SGD) and backpropagation.
The order of training data was shuffled and classes were balanced by sampling the same number of positive examples as negative examples per batch. Additionally, our \texttt{MolGridDataLayer} has the ability to randomly rotate and translate the input structures on-the-fly. This feature is controlled via data augmentation parameters specifying whether to randomly rotate structures and the maximum distance to randomly translate them. Enabling this data augmentation significantly improved training, as shown in Figure~\ref{depth3_train}.

The values for training hyperparameters were initially evaluated in ranges common for neural network training, and these values were verified to behave reasonably for our data.  In general, training parameters within conventional ranges converged to similar loss values, with the main difference being the number of iterations needed to converge.  The same parameters for the SGD solver (batch\_size=10, base\_lr=0.01, momentum=0.9), for learning rate decay (lr\_policy = inverse, power=1,  gamma=0.001), and for regularization (weight\_decay=0.001, dropout\_ratio=0.5) were used to train all models.
In all cases we manually verified that model training had qualitatively converged after 10,000 iterations.

\subsection{Model Evaluation}

The performance of trained CNN models were evaluated by 3-fold cross-validation for both the pose prediction and virtual screening tasks. To avoid evaluating models on targets similar to those in the training set, training and test folds were constructed by clustering data based on target families rather than individual targets. For the CSAR pose prediction training set, clusters were created using the 90\% sequence identity families provided by CSAR (i.e., protein targets with greater than 90\% sequence identity are always retained in the same fold to avoid testing on a target highly similar to one in the training set).
For the DUD-E virtual screening dataset, we created our own clusters of proteins using the hierarchical clustering module of \texttt{scipy} and ensured that proteins with greater than 80\% sequence identity were retained in the same fold.
 Receiver-operating characteristic (ROC) curves were generated for each model, plotting the true positive rate against the false positive rate. The performance metric was the area under the ROC curve (AUC), with AUC = 1 representing a perfect classifier and AUC = 0.5 being no better than chance.

\subsubsection{Independent Test Sets}

To control for any systematic bias in the training sets, we also chose to assess classification accuracy on several completely independent test sets. To evaluate pose prediction performance, we utilized the 2013 PDBbind core set\cite{pdbbind}.   The PDBbind database consists of high quality protein-ligand complexes with no unusual atomic features, such as uncommon elements. The core set is a representative, non-redundant subset of the database and is  composed of 195 protein-ligand complexes in 65 families. 

To assess virtual screening performance, we utilized two datasets created from assay results. One was generated from ChEMBL by Riniker and Landrum \cite{riniker2013}, following Heikamp and Bajorath \cite{heikamp2011}. They selected a set of 50 human targets from ChEMBL version 14. They chose actives that had at least 10 $\mu$M potency, had a molecular weight under 700 $g/mol$, and did not have metal ions. The actives were down sampled using the RDKit diversity picker to select the 100 most diverse compounds for each target. For each active, two decoys with a Dice similarity greater than 0.5 using a simple atom-count fingerprint (ECFC0) were randomly selected from the ZINC database to yield a total of 10,000 decoys that were shared across all targets. Our other virtual screening dataset is a subset of the maximum unbiased validation (MUV) dataset \cite{rohrer2009}, which is based on PubChem bioactivity data. MUV consists of assay data from 17 targets, each with 30 actives and 15000 decoys. Actives were selected from confirmatory screens and were chosen to be maximally spread based on simple descriptors and embedded in decoys. The decoys were selected from a primary screen for the same target. The MUV datasets were designed to avoid analog bias and artificial enrichment, which produce overly optimistic predictions of virtual screening performance.

To avoid artificially enhancing our performance on these test sets, we enforced a maximum similarity between targets included in the test sets and targets from DUD-E and CSAR used for training. We performed a global sequence alignment for all targets from the training and proposed test sets and removed any test targets that had more than 80\% sequence identity with a training target. We also performed ProBiS \cite{probis} structural alignment on the binding sites of all pairs of targets from the training and proposed test sets and rejected those for which a significant alignment was found using the default ProBiS parameters. Finally, since structural data were necessary for scoring, assay targets were only included if a crystal structure of a bound complex containing the target was available in the Protein Data Bank. This structure was used to generate docked poses at a known binding site. After these constraints were applied, the independent test sets consisted of  a 54 complex subset of the 2013 PDBbind core set, a 13 target subset of the Riniker and Landrum ChEMBL set, and a 9 target subset of the MUV set. 

For the pose prediction task, we re-docked ligands from the PDBbind core set with the settings \texttt{--seed 0 --exhaustiveness 50 --num\_modes 20} (the same settings used to generate poses for the CSAR training set). The resulting PDBbind core subset had 98 low RMSD ($<2${\AA}) out of 897 total poses. For the virtual screening task, the active and decoy sets were docked against an appropriate reference receptor using smina's default arguments for exhaustiveness and sampling. All generated poses were scored and the best score for each ligand was used to assess virtual screening performance. The resulting ChEMBL subset had 11,406 poses associated with 1,300 active compounds and 663,671 poses associated with 10,000 decoys. The resulting MUV subset had 1,913 poses associated with 270 active compounds and 1,177,989 poses associated with 135,000 decoys.

The ChEMBL and MUV test sets provide collections of actives and decoys associated with a target protein, but they do not provide crystal structures for the target. We only included targets with bound crystal structures available, and we used the bound ligand to identify the pocket into which to dock the assay's actives and decoys. Table ~\ref{testset_targets} shows the PDB accession code for the crystal structure we used for each target, the bound ligand associated with that structure, that ligand's experimental affinity for the target (if available), and the type of assay used to identify the actives and decoys. 

\begin{table}[tbph]
\begin{tabular}{| l | l | l | l | l | l |}
	\hline
	MUV ID & PDB ID & Ligand & $K_{i}$/$IC_{50}$ (nM) & Assay type \\ \hline \hline
	600 & 1yow & P0E & N/A & cell \\ 
	692 & 1yow & P0E & N/A & cell \\ 
	859 & 5cxv & 0HK & N/A & cell \\ 
	852 & 4xe4 & NAG & N/A & biochemical \\ 
	548 & 3poo & S69 & N/A & biochemical \\ 
	832 & 1au8 & 0H8 & N/A & biochemical \\ 
	689 & 2y6o & 1N1 & 25 & biochemical \\ 
	846 & 5exm & 5ST & N/A & biochemical \\ 
	466 & 3v2y & ML5 & 18-77 & cell \\ \hline
\end{tabular}
\par
\vspace*{1 cm}
\begin{tabular}{| l | l | l | l | l | l |}
	\hline
	ChEMBL ID & PDB ID & Ligand & $K_{i}$/$IC_{50}$ (nM) & Assay type \\ \hline \hline
	10752 & 4kik & KSA & N/A & biochemical \\
	11359 & 1mkd & ZAR & 160 & biochemical \\ 
	12209 & 4ht2 & V50 & 150-290 & biochemical \\ 
	28 & 1hvy & D16 & 290 & biochemical \\ 
	276 & 2qyk & NPV & 1-88 & biochemical \\ 
	10498 & 2xu1 & 424 & 22 & biochemical \\ 
	11534 & 1ms6 & BLN & 0.3 & biochemical \\ 
	10378 & 1csb & EP0 & N/A & biochemical \\ 
	219 & 4daj & 0HK & N/A & biochemical \\ 
	11279 & 3ks9 & Z99 & 6800 & biochemical \\ 
	12968 & 4s0v & SUV & 0.35-12 & biochemical \\ 
	20014 & 1mq4 & ADP & N/A & biochemical \\ 
	11631 & 3v2y & ML5 & 18-77 & biochemical \\ 
	18061 & 5ek0 & 5P2 & N/A & biochemical \\ 
	12670 & 4xuf & P30 & 1.3-8.8 & biochemical \\ \hline
\end{tabular}
\caption{\label{testset_targets}Information about the PDB structures chosen to provide structural information for the selected test set targets. The PDB accession code, crystal ligand, affinity of the crystal ligand for the protein, and assay type from which the test set actives and decoys were derived are shown.}
\end{table}

\subsection{Optimization}

An initial CNN architecture was constructed using simple guidelines in order to limit parameterization and serve as a starting point for optimization. The preliminary model architecture consisted of five  3x3x3 convolutional layers with rectified linear activation units alternating with max pooling layers. The number of filters in each convolutional layer was doubled from the previous one so that the width of the network increased as the spatial dimensionality decreased. Following the alternating convolution and pooling layers was a single fully connected layer with two outputs and a softmax layer for binary classification.

The various parameters of the neural network model were tuned to train the most accurate model with respect to the CSAR pose prediction test set.  The CSAR set was chosen as its smaller size made iterative model optimization more practical. Model optimization was performed by systematically modifying a reference model.  A single parameter was varied and the resulting training times and accuracies computed. After all parameters were tested, the changes resulting in the best gain of accuracy and similar or reduced training time were combined to create a new reference model. This process was repeated until the model's accuracy no longer increased.  Several model parameters were explored.

\paragraph{Atom Types}
In addition to the default smina atom types, we evaluated two simpler atom typing schemes: element-only and ligand/receptor only.  Unlike smina atom types, which include aromaticity and protonation state information, element-only types only record the element, although we still provide distinct types for receptor and ligand atoms.
With ligand/receptor only types, there are only two types (corresponding to two ``channels'' in the input 3D image): ligand atoms and receptor atoms. 

\paragraph{Occupancy Type}  In addition to a smoothed Gaussian distribution of atom density, we also evaluated a Boolean representation, where grid point values are one if they overlap an atom and zero otherwise.  Unlike with the Gaussian scheme, in the Boolean representation individual grid point values provide no indication of the distance of the grid point from the atom center.

\paragraph{Atomic Radius Multiplier}  By default, we extend atom densities beyond the van der Waals radius by a multiple of 1.5 (e.g., if the atomic radius is 1.0, the atom density decays to zero at 1.5).  Additionally, we evaluated multiples of 1.0, 1.25, 1.75, and 2.0.  With larger multiples, a single grid point contains more information about the local neighborhood.

\paragraph{Resolution}  The default grid resolution is 0.5{\AA} resulting in 48$^3$ grid points. We also evaluated higher (0.25{\AA}) and lower (0.75, 1.0, and 1.5{\AA}) resolution grids.

\paragraph{Layer Width}  In our initial reference model, the first convolutional layer generates 128 feature maps, and each successive layer doubles the number of feature maps after halving the dimensions of the maps with a pooling layer.  We also evaluate models that double, half, and quarter the width of these layers.  Wider layers allow for a more expressive model, but at the cost of more computation.

\paragraph{Model Depth} Our initial model contained 5 convolution layers. We also evaluate models with more (up to 8) and fewer (as little as 1) convolution layers.  More layers allow for a more expressive model, but take longer to process and increase the risk of suffering from vanishing gradients, which inhibit convergence \cite{bengio1994learning}.

\paragraph{Pooling Type}
Pooling layers reduce the size of their inputs by propagating a single value for each window (or kernel) of the input.  The propagated value can either be the maximum value or the average value of the kernel and the kernel size can be varied.  In our initial model we use max pooling with a kernel size of 2. We additionally evaluate average pooling and kernels of size 4.

\paragraph{Fully Connected Layer} After a series of convolution and pooling layers, a traditional fully connected layer reduces the final feature maps to two outputs. Our initial model contains a single fully connected layer.  Additionally, we evaluate alternative models with a single hidden layer with anywhere from  6 to 50 nodes.
More expressive fully connected layers allow the model to arbitrarily combine the spatial features generated by the convolution layers to generate the final prediction.

\begin{figure}[tbp]
	\includegraphics[width = 450px]{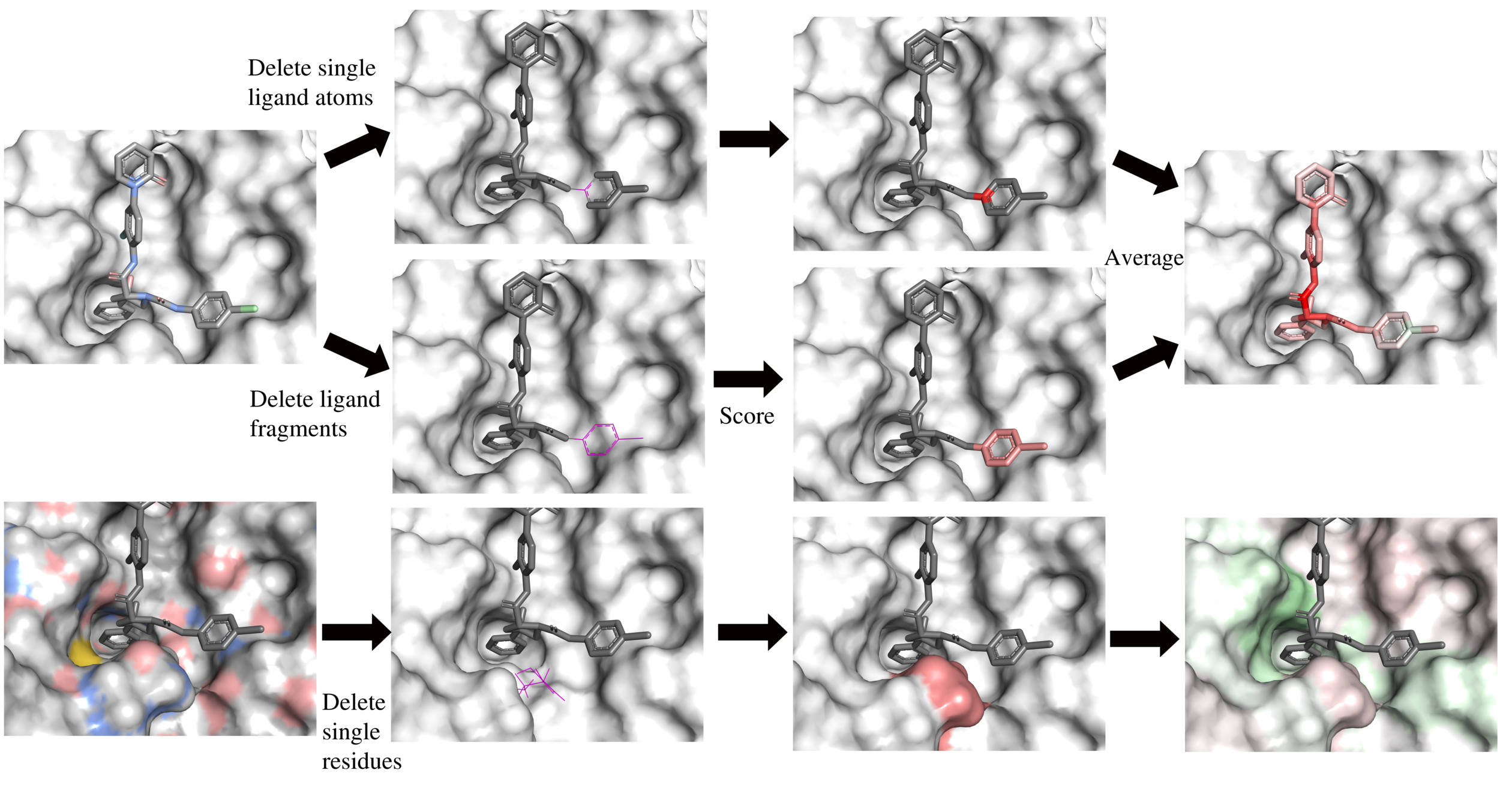}
    \caption{\label{vizfig}Visualization algorithm. In the ligand, atoms are removed individually or as fragments and each modified molecule is scored. The assigned color is the difference between the unmodified protein-ligand score and the score with the removed atom. The protein is treated similarly, but whole residues are removed.
    Positive score differences indicate a positive contribution by the atom to the
overall score and are colored green, with the intensity depending on the
magnitude of difference. Red represented negative score differences.
    }
\end{figure}

\subsection{Visualization}

In order to better understand the features that the neural network learns, we
implemented a visualization algorithm based on masking \cite{szegedy2013NIPS}. 
In image recognition masking, pixels are systematically masked out and the image is
reclassified in order to get a ``heat map'' of important areas.
The visualization algorithm is illustrated in
Figure~\ref{vizfig}. Atoms are colored by relative contribution to the total neural network score as 
determined by removing the atom and rescoring the complex.

Atoms are removed either one at a time, or as part of larger fragments.
The individual and fragment removals of atoms differ significantly enough that
an average of both scores is computed. The individual removals produce sharper
contrasts between ``good'' and ``bad'', compared to a more gradual effect in
the fragment removals. The combination of the two methods provides a broader
representation of how the model interprets functional groups, while maintaining
any significant individual atom scores. 

In order to reduce computational load, removals were carried out on whole
residues of the protein at a time. This provided enough information to assess
spatial relationships between protein and ligand, which is a key goal of
visualization. 

\section{Results}

Our systematic optimization of network and training parameters successfully improved the performance of the CNN models in clustered cross-validation while revealing the importance, or lack thereof, of various choices of parameters.  We evaluated the optimized network architecture for performance in pose prediction, virtual screening, and affinity prediction, while also considering the importance of the training set used to create the model. 

\begin{figure}[tbp]
\includegraphics[width=0.8\linewidth]{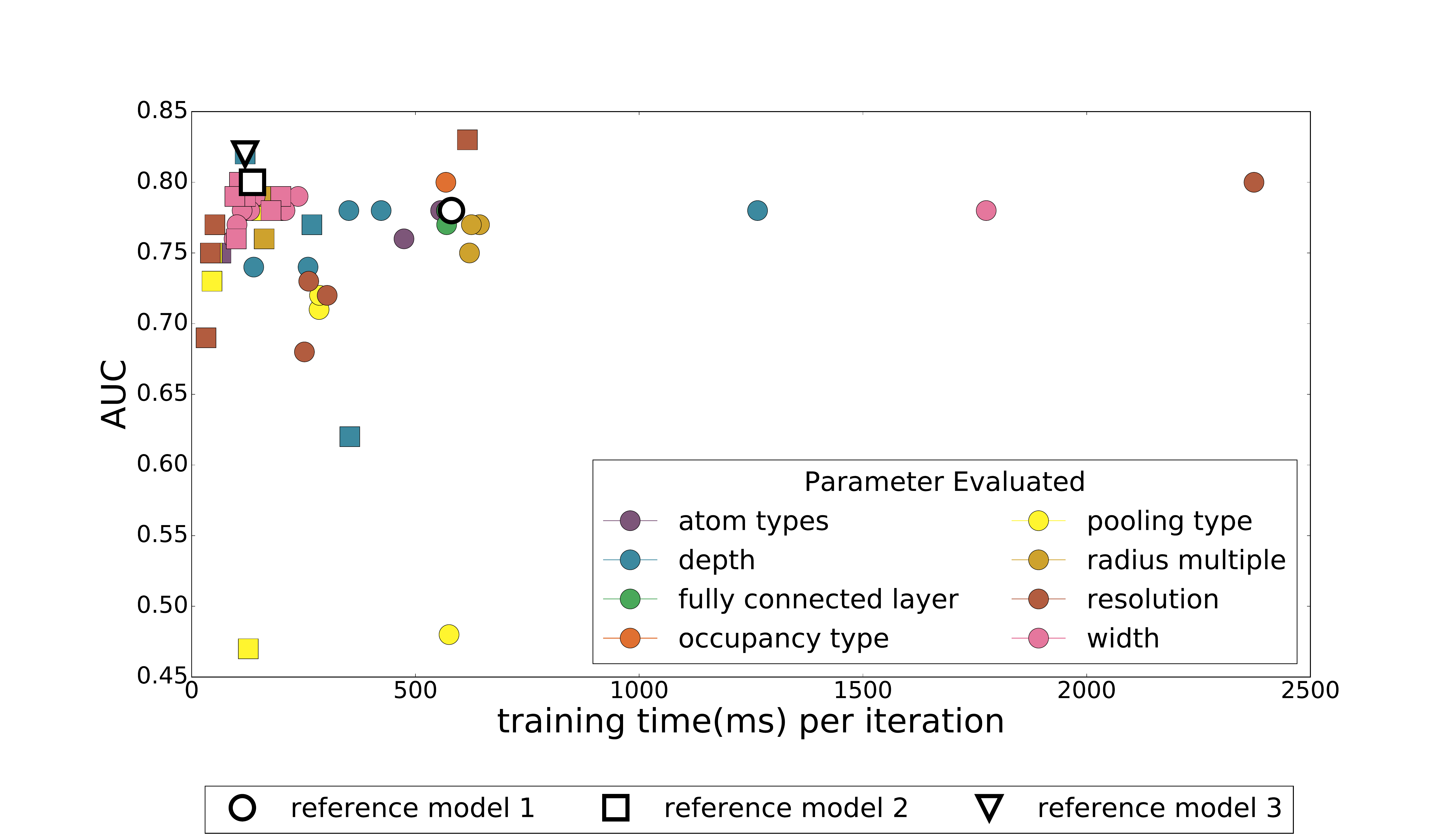}
\caption[]{\label{timevsauc} The training time and average cross-validation AUC of various models created by systematically varying parameters.  Marker shape indicates iteration of optimization and the color what parameter was varied.}
\end{figure}

\begin{figure}[tbp]
\includegraphics[width=0.5\linewidth]{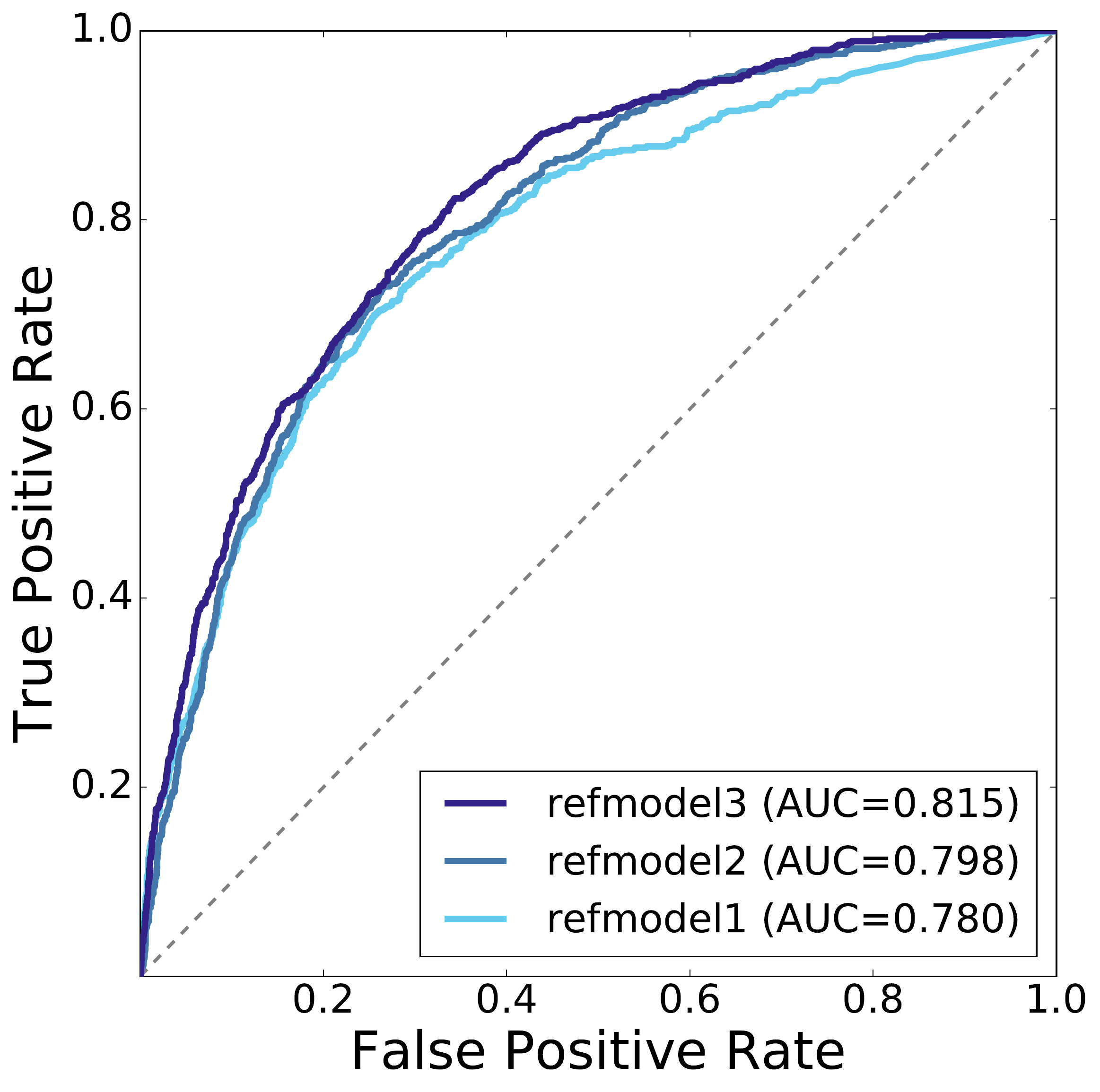}
\caption[]{\label{refmodelrocs} The ROC curves for the three reference models used during model optimization.}
\end{figure}

\begin{figure}[tbp]
\includegraphics[width=0.2\linewidth]{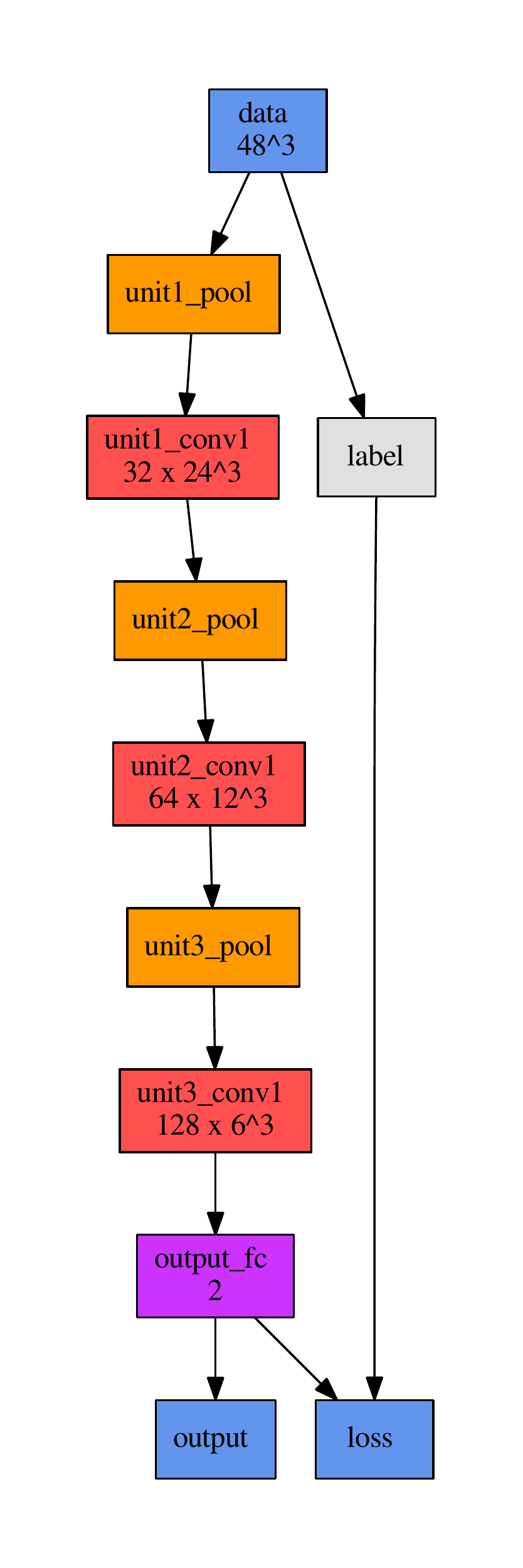}
\caption[]{\label{depth3} The network architecture of our final model.}
\end{figure}

\subsection{Optimization}

Two rounds of model optimization were performed.  In each round, parameters of a reference model were individually varied.  For each parameter type, the best parameter was used to define the reference model of the next iteration.  Each iteration both increased the cross-validation AUC and decreased the training time of the model. The results obtained in the first two iterations are shown in Figure~\ref{timevsauc}.  A third iteration did not result in further improvements (data not shown). The initial reference model had an AUC of 0.78 and a training time of 580ms per an iteration, and the final model increased to an AUC of 0.82 with a training time of 120ms per an iteration. The ROC curves for all three models are shown in Figure~\ref{refmodelrocs}. 

Based on the first iteration of parameter sensitivity analysis, the second reference model  reduced the depth from five to four convolutional layers and quartered the widths of these layers. After another round of optimization, the final reference model further reduced the depth to three convolutional layers. The final optimized network architecture is shown in Figure~\ref{depth3}.
Since parameters were varied individually in each optimization iteration, we can assess the relative importance of each parameter class on the overall model performance.

\paragraph{Atom Types}  The best AUCs are achieved using smina atom types.  However, simpler atom types are remarkably competitive, with at most a 0.05 reduction in AUC for the binary protein/ligand atom typing in the second iteration of optimization.  This is consistent with previous findings with empirical scoring functions where purely steric terms were found to be the dominant terms of the scoring function \cite{sminapaper,Novikov}.  Additionally, although the overall AUCs were similar, smina and element-only atom types result in better early enrichment (the initial slope of the ROC curve is steeper).

\paragraph{Occupancy Type}  Interestingly, changing the atom density representation from the more informative Gaussian to a simple Boolean did not reduce the AUC.  The models do not seem to need the additional distance information provided by a Gaussian atomic density.

\paragraph{Atomic Radius Multiplier}  The default radius multiplier of 1.5 provided the best AUC, although other multipliers were nearly equivalent with all but the 2.0 multiplier within 0.01 of the reference AUC.  

\paragraph{Resolution} Predictive performance correlates with resolution, with the highest resolution (0.25{\AA}) achieving an AUC more than 0.1 greater than the lowest (1.5{\AA}).  However, we decided against using higher resolution grids since the small increase in AUC (0.02) in increasing the resolution from 0.5{\AA} to 0.25{\AA} was accompanied by a more than 4X increase in training time.  

\paragraph{Layer Width}  We found that increasing the width of the layers resulted in significant increases in training time, but slight decreases in predictive performance, possibly due to overfitting.  Reducing the width improved both the AUC and training time up to a limit.  In our final model, the first convolutional layer generates 32 feature maps; reducing this number further hurts predictive performance.

\paragraph{Model Depth} Model depth behaved similarly to the layer width parameter.  Our initial model topology was needlessly expressive, and by reducing the depth (ultimately to only three convolutional layers), we improved both training time and predictive performance, likely by reducing the amount of overfitting. 

\paragraph{Pooling Type} Somewhat surprisingly, the use of average pooling instead of max pooling obliterated predictive performance and prevented the model from learning.  Alternative kernel sizes did not improve the AUC.

\paragraph{Fully Connected Layer} 
Modifications to the final fully connected layer had no discernible effects on predictive performance or training time, suggesting most of the learning is taking place in the convolutional layers.

The final optimized model architecture was used to train and evaluate pose prediction, virtual screening, and affinity prediction performance.  It is available at \url{https://github.com/gnina/models}.

\subsection{Pose Prediction}

Pose prediction assesses the ability of a scoring function to distinguish between low RMSD and high RMSD poses of the same compound.  We assess pose prediction performance both in terms of inter-target ranking and intra-target ranking.  With inter-target ranking, which is most similar to the training protocol, all poses across all targets are ranked to generate a ROC curve.  Intra-target ranking better represents the typical docking scenario, and the goal is to select the the lowest RMSD pose among poses generated for each individual target.  A scoring function can do well in intra-target ranking even if the low RMSD pose has a poor score as long as all other poses for that target have worse scores.

\begin{figure}[tbp]
\includegraphics[width=0.5\linewidth]{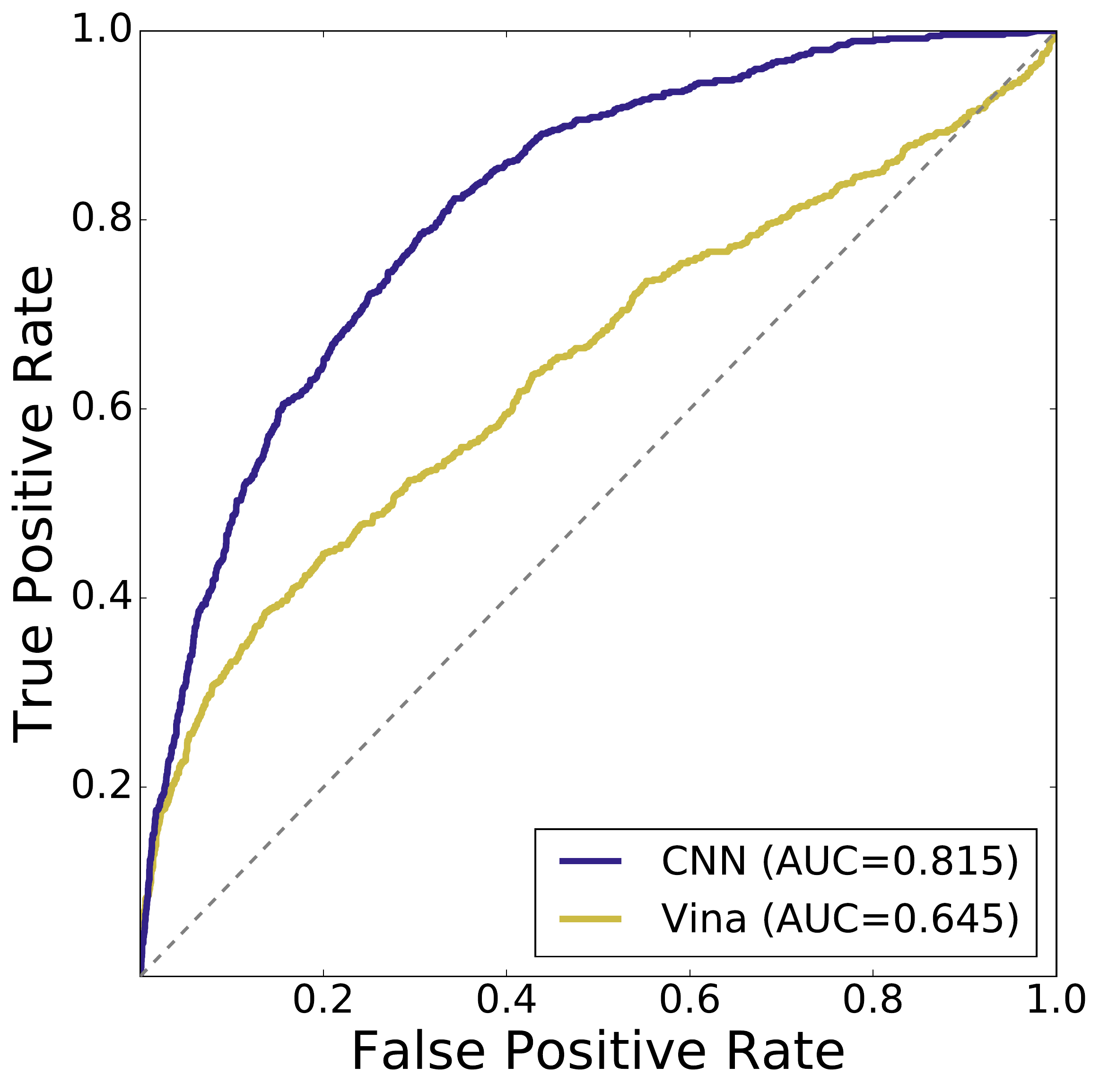}
\caption[]{\label{depth3_all_roc} Inter-target cross-validated ROC curve of CNN scoring method compared to Autodock Vina on the CSAR pose prediction data set. The CNN performs better at classifying generated poses as low or high RMSD across targets.}
\end{figure}

\begin{figure}[tpb]
\includegraphics[width=0.6\linewidth]{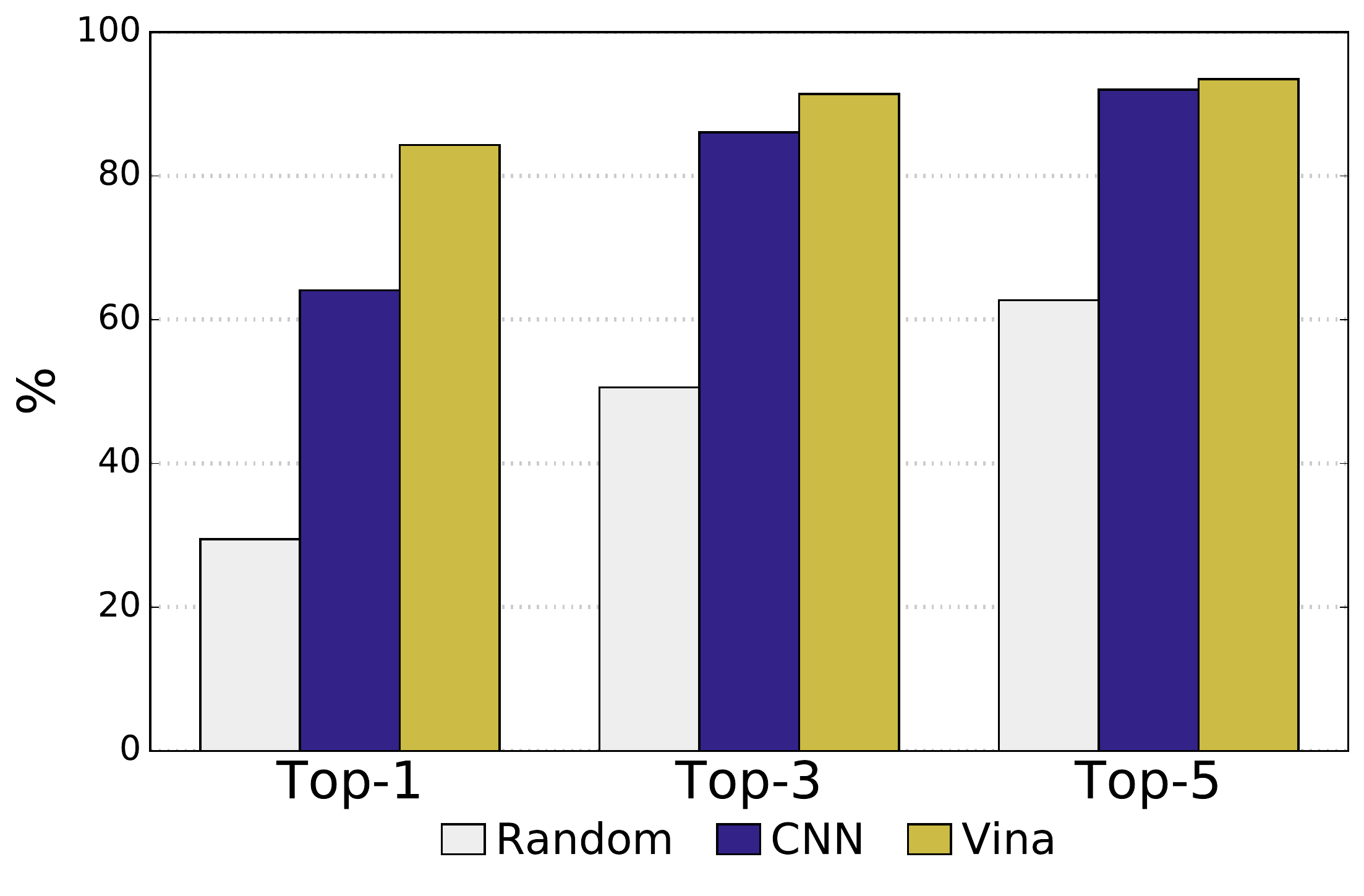}
\caption[]{\label{ranking_bar} Intra-target pose ranking. The percent of targets with a low RMSD pose ranked as the top one, three, or five poses is shown. Vina and CNN have similar recovery rates among the top-5 ranked poses, but Vina more often ranks a low RMSD pose as the top-1 ranked pose.}
\end{figure}

\begin{figure}[tbp]
\includegraphics[width=1.0\linewidth]{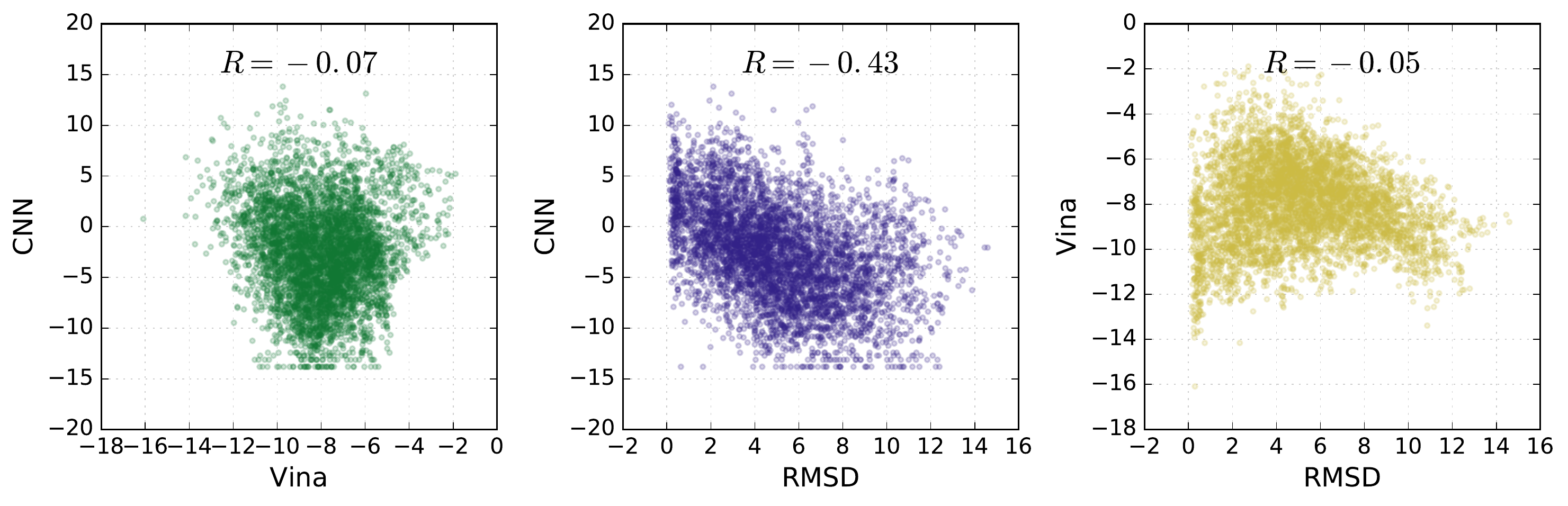}
\caption[]{\label{depth3_all_cor} Pearson correlation between Vina score, CNN score, and RMSD from crystal pose. A logit transformation has been applied to the CNN score, mapping it from probability to linear space, in order to more easily see the relationship. All generated poses are shown, including those with RMSD between 2 and 4 which were omitted from training.}
\end{figure}

The CNN model performed substantially better than the Autodock Vina scoring function in its ability to perform inter-target ranking of CSAR poses as shown by the cross-validation results in Figure~\ref{depth3_all_roc}.
The CNN model achieves an AUC of 0.815 while the Vina scoring function has an AUC of 0.645.

In intra-target ranking, the CNN model performed substantially worse than Autodock Vina, as shown in Figure~\ref{ranking_bar}.  The Autodock Vina scoring function is parameterized to excel at redocking \cite{Trott2009,sminapaper} and correctly identifies a low RMSD pose as the top ranked pose for the given target for 84\% of the targets compared to 64\% with the CNN model.  When the top 5 poses are considered, the difference between Vina and the CNN model shrinks with Vina exhibiting a success rate of 93\% and the CNN model 92\%.  As pose selection performance is dependent on the range of poses that are selected from (e.g., some targets have highly rigid ligands in tightly constrained pockets resulting in nearly all low RMSD poses), we also show the results of random selection in Figure~\ref{ranking_bar}.  Both methods are substantially better than random.

The correlations between pose RMSD and scores are shown in Figure~\ref{depth3_all_cor}.  The CNN scores weakly correlate with RMSD, with higher RMSD poses exhibiting lower scores as expected (a more positive CNN score is more favorable).  Vina scores do not correlate with RMSD, although there is a noticeable ``funnel'' shape due to the best scoring poses having very low RMSDs.  Interestingly, there is no correlation between CNN scores and Vina scores, indicating that they use different criteria to rank poses.

\subsection{Virtual Screening}

Structure-based virtual screening assesses the ability of a scoring function to distinguish between active and inactive compounds using docked structures.  In assessing virtual screening, we consider both the case where the CNN model ranks only the top-ranked (by Vina) docked pose of each ligand (single-pose prediction) and the case where the CNN model selects from all available docked poses of the ligand (multi-pose prediction).

\begin{figure}[tbp]
\includegraphics[width=0.4\linewidth]{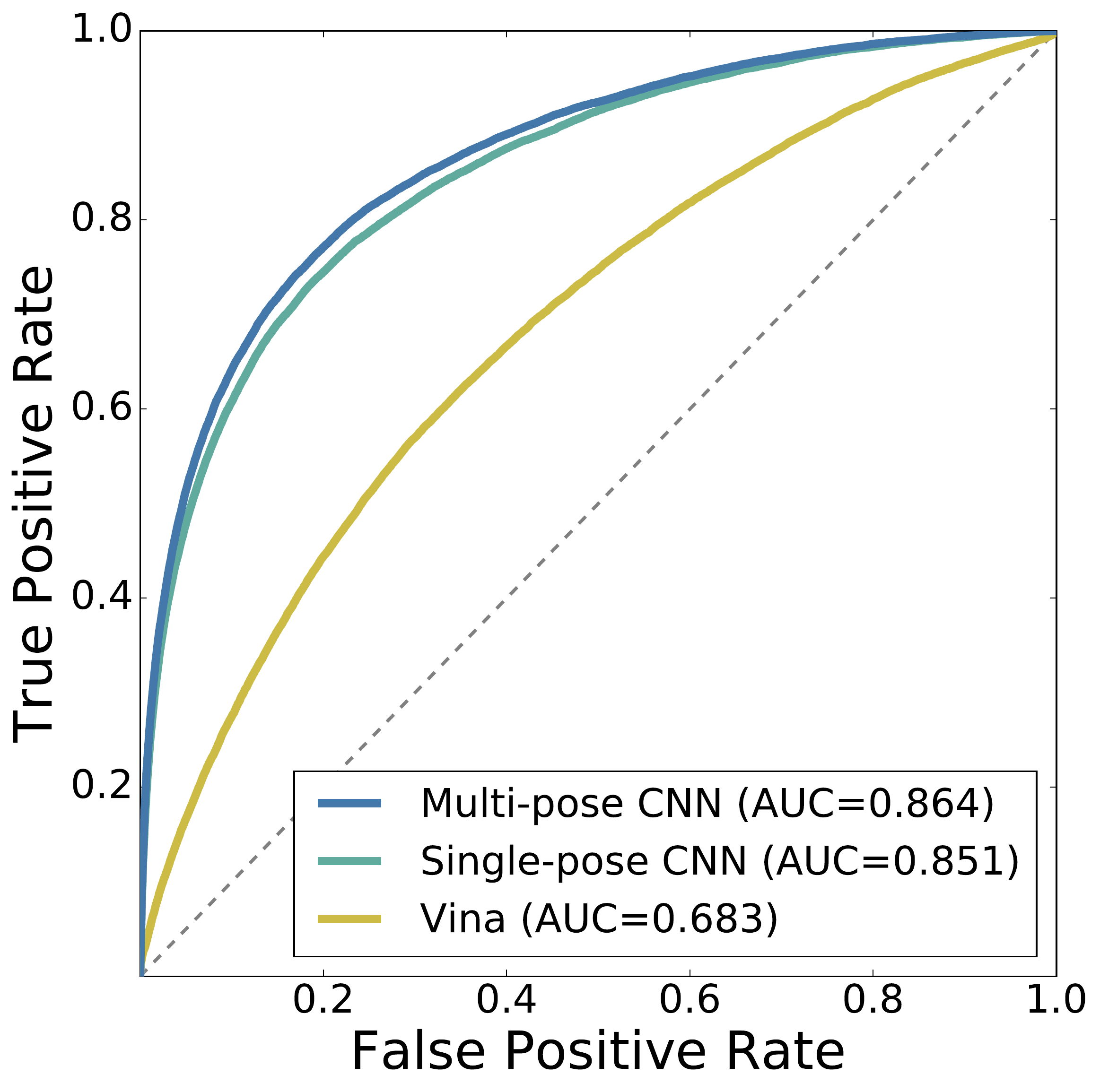}
\caption[]{\label{duderoc} ROC curves for cross-validation virtual screening performance across the full DUD-E benchmark. Single-pose scoring distinguishes between active and inactive compounds using the top ranked pose identified by Vina while multi-pose scoring selects among all docked poses using the maximum CNN score.}
\end{figure}

\begin{figure}[tbp]
\includegraphics[width=1.0\linewidth]{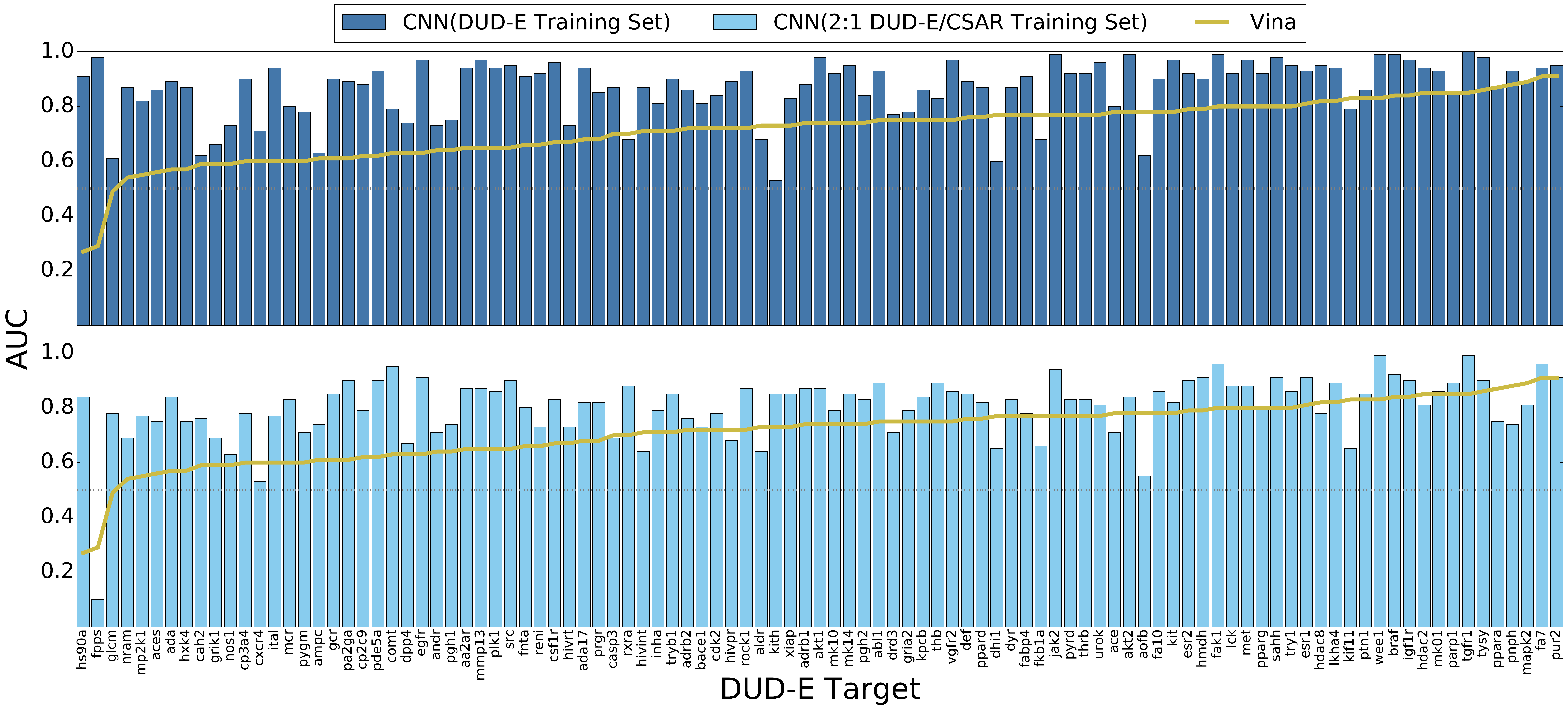}
\caption[]{\label{bytarget} Cross-validation performance of CNN models on the DUD-E virtual screening benchmark compared to the Vina scoring function. Targets are sorted by performance with Vina. Identical sets of docked poses were ranked.  The score of the top ranked pose of each ligand is used to predict activity (multi-pose scoring).  CNN models trained only on DUD-E training data perform best, outperforming Vina in 90\% of the targets.  Models trained using a mix of DUD-E and CSAR data also perform well, achieving better AUCs than Vina in 81\% of the targets.}
\end{figure}

Overall cross-validation results for the entire DUD-E benchmark are shown in Figure~\ref{duderoc} and Table~\ref{dudeaucs}.  Even using the exact same poses (single-pose scoring), CNN scoring substantially outperforms Vina with an AUC of 0.85 versus 0.68. Multi-pose scoring does slightly better with an AUC of 0.86. On a per-target basis, CNN scoring outperforms Vina scoring for 90\% of the DUD-E targets, as shown in Figure~\ref{bytarget}.

\begin{figure}[tbp]
\includegraphics[width=0.8\linewidth]{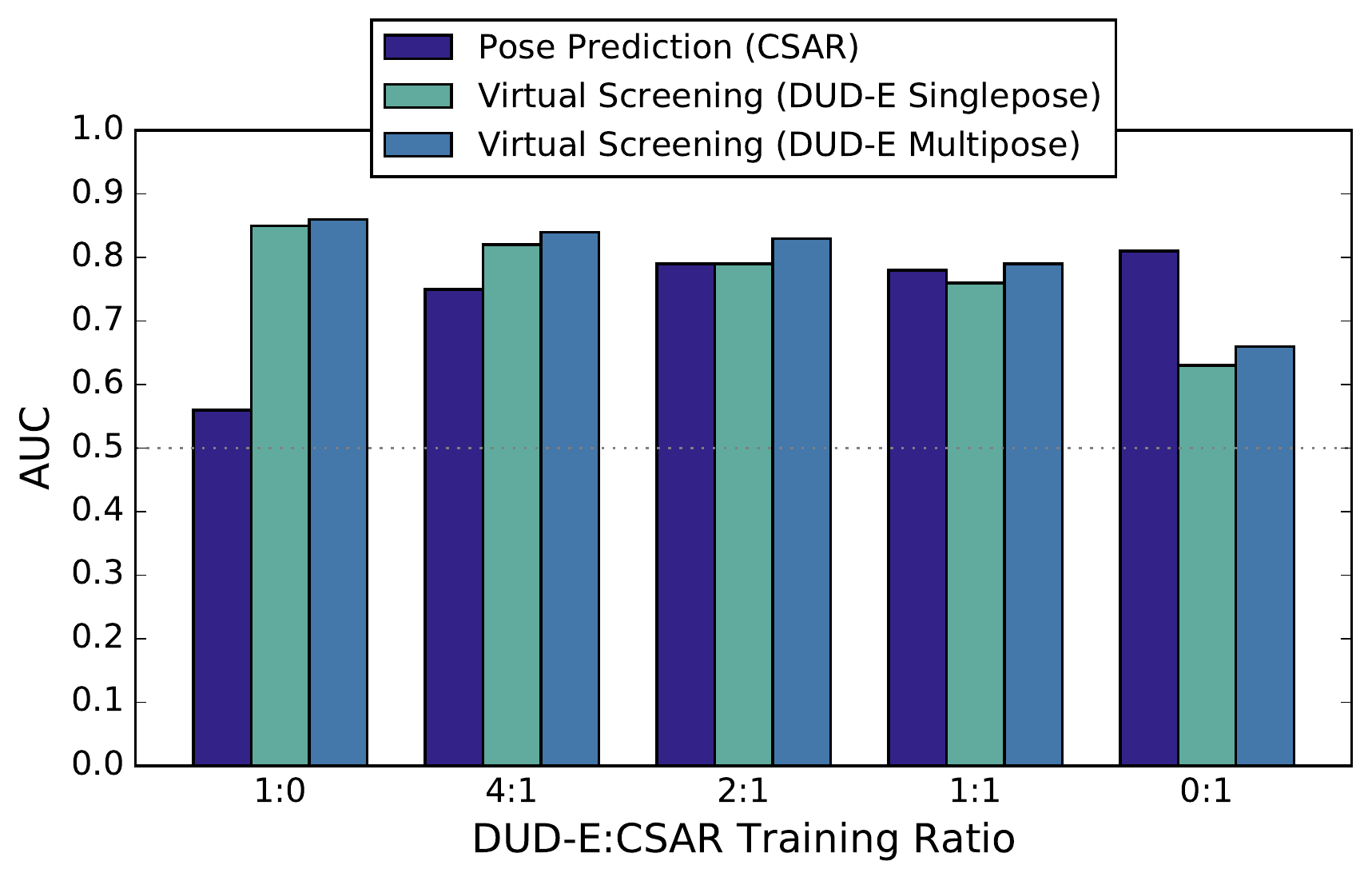}
\caption[]{\label{mixed_data_bar} The cross-validation performance of the CNN model when trained with different ratios of CSAR and DUD-E data and evaluated in terms of pose prediction (CSAR) and virtual screening (DUD-E).}
\end{figure}

\begin{table}[tbph]
\scriptsize
\begin{tabular}{r|c|c|c}
Target & Vina & DUD-E& 2:1 D/C  \\ \hline \hline
aa2ar & 0.65 & 0.94 & 0.87\\
abl1 & 0.75 & 0.93 & 0.89\\
ace & 0.78 & 0.80 & 0.71\\
aces & 0.56 & 0.86 & 0.75\\
ada & 0.57 & 0.89 & 0.84\\
ada17 & 0.68 & 0.94 & 0.82\\
adrb1 & 0.74 & 0.88 & 0.87\\
adrb2 & 0.72 & 0.86 & 0.76\\
akt1 & 0.74 & 0.98 & 0.87\\
akt2 & 0.78 & 0.99 & 0.84\\
aldr & 0.73 & 0.68 & 0.64\\
ampc & 0.61 & 0.63 & 0.74\\
andr & 0.64 & 0.73 & 0.71\\
aofb & 0.78 & 0.62 & 0.55\\
bace1 & 0.72 & 0.81 & 0.73\\
braf & 0.84 & 0.99 & 0.92\\
cah2 & 0.59 & 0.62 & 0.76\\
casp3 & 0.70 & 0.87 & 0.69\\
cdk2 & 0.72 & 0.84 & 0.78\\
comt & 0.63 & 0.79 & 0.95\\
cp2c9 & 0.62 & 0.88 & 0.79\\
cp3a4 & 0.60 & 0.90 & 0.78\\
csf1r & 0.67 & 0.96 & 0.83\\
cxcr4 & 0.60 & 0.71 & 0.53\\
def & 0.76 & 0.89 & 0.85\\
dhi1 & 0.77 & 0.60 & 0.65\\
dpp4 & 0.63 & 0.74 & 0.67\\
drd3 & 0.75 & 0.77 & 0.71\\
dyr & 0.77 & 0.87 & 0.83\\
egfr & 0.63 & 0.97 & 0.91\\
esr1 & 0.81 & 0.93 & 0.91\\
esr2 & 0.79 & 0.92 & 0.90\\
fa10 & 0.78 & 0.90 & 0.86\\
fa7 & 0.91 & 0.94 & 0.96
\end{tabular}
\begin{tabular}{|r|c|c|c|}
Target & Vina & DUD-E& 2:1 D/C  \\ \hline \hline
fabp4 & 0.77 & 0.91 & 0.78\\
fak1 & 0.80 & 0.99 & 0.96\\
fkb1a & 0.77 & 0.68 & 0.66\\
fnta & 0.66 & 0.91 & 0.80\\
fpps & 0.29 & 0.98 & 0.10\\
gcr & 0.61 & 0.90 & 0.85\\
glcm & 0.49 & 0.61 & 0.78\\
gria2 & 0.75 & 0.78 & 0.79\\
grik1 & 0.59 & 0.66 & 0.69\\
hdac2 & 0.85 & 0.94 & 0.81\\
hdac8 & 0.82 & 0.95 & 0.78\\
hivint & 0.71 & 0.87 & 0.64\\
hivpr & 0.72 & 0.89 & 0.68\\
hivrt & 0.67 & 0.73 & 0.73\\
hmdh & 0.79 & 0.90 & 0.91\\
hs90a & 0.27 & 0.91 & 0.84\\
hxk4 & 0.57 & 0.87 & 0.75\\
igf1r & 0.84 & 0.97 & 0.90\\
inha & 0.71 & 0.81 & 0.79\\
ital & 0.60 & 0.94 & 0.77\\
jak2 & 0.77 & 0.99 & 0.94\\
kif11 & 0.83 & 0.79 & 0.65\\
kit & 0.78 & 0.97 & 0.82\\
kith & 0.73 & 0.53 & 0.85\\
kpcb & 0.75 & 0.86 & 0.84\\
lck & 0.80 & 0.92 & 0.88\\
lkha4 & 0.82 & 0.94 & 0.89\\
mapk2 & 0.89 & 0.89 & 0.81\\
mcr & 0.60 & 0.80 & 0.83\\
met & 0.80 & 0.97 & 0.88\\
mk01 & 0.85 & 0.93 & 0.86\\
mk10 & 0.74 & 0.92 & 0.79\\
mk14 & 0.74 & 0.95 & 0.85\\
mmp13 & 0.65 & 0.97 & 0.87\\
\end{tabular}
\begin{tabular}{|r|c|c|c}
Target & Vina & DUD-E& 2:1 D/C  \\ \hline \hline
mp2k1 & 0.55 & 0.82 & 0.77\\
nos1 & 0.59 & 0.73 & 0.63\\
nram & 0.54 & 0.87 & 0.69\\
pa2ga & 0.61 & 0.89 & 0.90\\
parp1 & 0.85 & 0.85 & 0.89\\
pde5a & 0.62 & 0.93 & 0.90\\
pgh1 & 0.64 & 0.75 & 0.74\\
pgh2 & 0.74 & 0.84 & 0.83\\
plk1 & 0.65 & 0.94 & 0.86\\
pnph & 0.88 & 0.93 & 0.74\\
ppara & 0.87 & 0.87 & 0.75\\
ppard & 0.76 & 0.87 & 0.82\\
pparg & 0.80 & 0.92 & 0.80\\
prgr & 0.68 & 0.85 & 0.82\\
ptn1 & 0.83 & 0.86 & 0.85\\
pur2 & 0.91 & 0.95 & 0.91\\
pygm & 0.60 & 0.78 & 0.71\\
pyrd & 0.77 & 0.92 & 0.83\\
reni & 0.66 & 0.92 & 0.73\\
rock1 & 0.72 & 0.93 & 0.87\\
rxra & 0.70 & 0.68 & 0.88\\
sahh & 0.80 & 0.98 & 0.91\\
src & 0.65 & 0.95 & 0.90\\
tgfr1 & 0.85 & 1.00 & 0.99\\
thb & 0.75 & 0.83 & 0.89\\
thrb & 0.77 & 0.92 & 0.83\\
try1 & 0.80 & 0.95 & 0.86\\
tryb1 & 0.71 & 0.90 & 0.85\\
tysy & 0.86 & 0.98 & 0.90\\
urok & 0.77 & 0.96 & 0.81\\
vgfr2 & 0.75 & 0.97 & 0.86\\
wee1 & 0.83 & 0.99 & 0.99\\
xiap & 0.73 & 0.83 & 0.85\\
 & & &
\end{tabular}
\caption{\label{dudeaucs} Cross-validation DUD-E AUCs for Vina and CNN models trained using either only DUD-E docked poses or a combination, at a 2:1 ratio, of DUD-E poses and CSAR poses.}
\end{table}

\subsection{Combined Training}

CNN models trained on one kind of data do not generalize particularly well to another. For example, as shown in Figure~\ref{mixed_data_bar}, a CNN model trained exclusively on DUD-E data achieves a cross-validation AUC of 0.56 in CSAR pose prediction.  This is not unexpected as the DUD-E training data consists of noisy, likely inaccurate, docked poses. A CNN model trained on this data will be less sensitive to changes in ligand pose.  In the other direction, training on CSAR data resulted in a cross-validation AUC of 0.66 at the virtual screening task.  However, as shown in Figure~\ref{mixed_data_bar}, combining CSAR and DUD-E training data results in models that perform nearly as well as single-task trained models.  At a ratio 2:1 DUD-E to CSAR (for every two virtual screening training examples from DUD-E, one pose prediction example from CSAR is included during training), the resulting CNN model exhibits an AUC of 0.79 at pose prediction and an AUC of 0.83 at virtual screening.  The inclusion of pose prediction training data accentuates the difference between single-pose and multi-pose DUD-E evaluation (e.g., 0.79 vs 0.83 at a 2:1 ratio), suggesting that such data allows the CNN model to select more accurate poses.

Although a combined training set results in a minimal reduction in overall AUC for DUD-E, on a per-target basis, shown in Figure~\ref{bytarget}, there is a more significant reduction in performance, with only 81\% of targets performing better than Vina, compared with 90\% with a DUD-E-only training set.  In a few cases, the difference is dramatic, such as with the fpps (farnesyl diphosphate synthase) target which goes from a 0.98 AUC with the DUD-E-only training set to a 0.10 AUC with the combined 2:1 training set.  This target is also a challenge for the Vina scoring function, which also achieves a worse-than-random 0.29 AUC, suggesting that the generated poses may be highly inaccurate.  

\begin{figure}
\centering
\subcaptionbox[]{\label{fppsatom}}[0.25\linewidth]{
\includegraphics[width=0.25\linewidth]{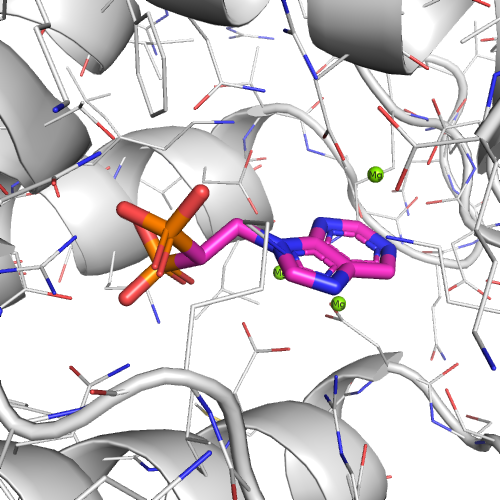}
}
\subcaptionbox[]{\label{fppscolor}}[0.25\linewidth]{
\includegraphics[width=0.25\linewidth]{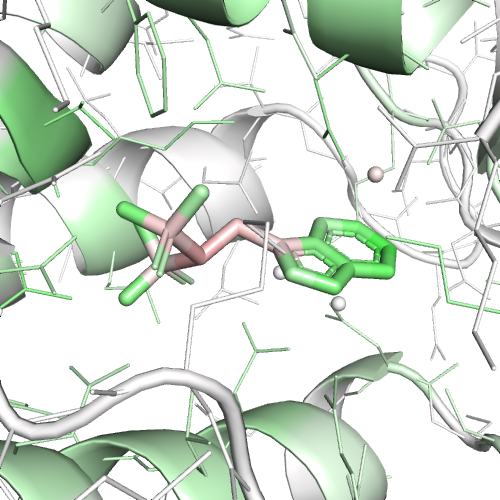}
}
\caption{\label{fpps} \subref{fppsatom} The top ranked pose by Vina of the CHEMBL457424 ligand of the fpps DUD-E target. \subref{fppscolor} Visualization of a CNN model trained using only DUD-E training data. The pose is scored highly due to the polar parts of the structure regardless of the orientation of the ligand.}
\end{figure}

An example ligand from the fpps target is CHEMBL457424, which the DUD-E model scores as 0.99 but the combined model scores as 0.01.  The DUD-E model is completely pose insensitive - all poses of this ligand score similarly despite large differences in RMSD.  The pose selected with Vina is shown visualized with the DUD-E model in Figure~\ref{fpps}. This pose is most likely incorrect; based on the 3ZOU crystal structure, the bisphosphonate group should chelate with the magnesium ions.  The DUD-E model highlights the polar and aromatic parts of the molecule and disfavors the apolar parts.  It also highlights the polar residues of the binding site.  It is possible that the DUD-E-only model is simply ranking polar molecules highly, having recognized the highly polar binding site.  Furthermore, all the actives associated with this target in the DUD-E benchmark contain a bisphosphonate group, whereas fewer than 1\% of the decoy compounds even contain phosphorous.  A scoring function that favors this group regardless of the 3D interaction structure will do exceptionally well in scoring these actives.
 When pose quality is incorporated into the training of the model, as with the combined model, erroneous poses are penalized and non-structural properties, such as polarity, play a less dominant role.  Similar trade-offs between learning non-structural cheminformatic information and enforcing structural constraints likely explain the difference in performance between the DUD-E and combined models.

\subsection{Independent Test Sets}

To evaluate CNN scoring performance on our independent test sets, we trained three models using all folds of the available training data: a pose prediction model trained only on CSAR data, a virtual screening model trained only on DUD-E data, and a combined model trained on DUD-E and CSAR data at a 2:1 ratio.

\begin{figure}[tbp]
\includegraphics[width=0.5\linewidth]{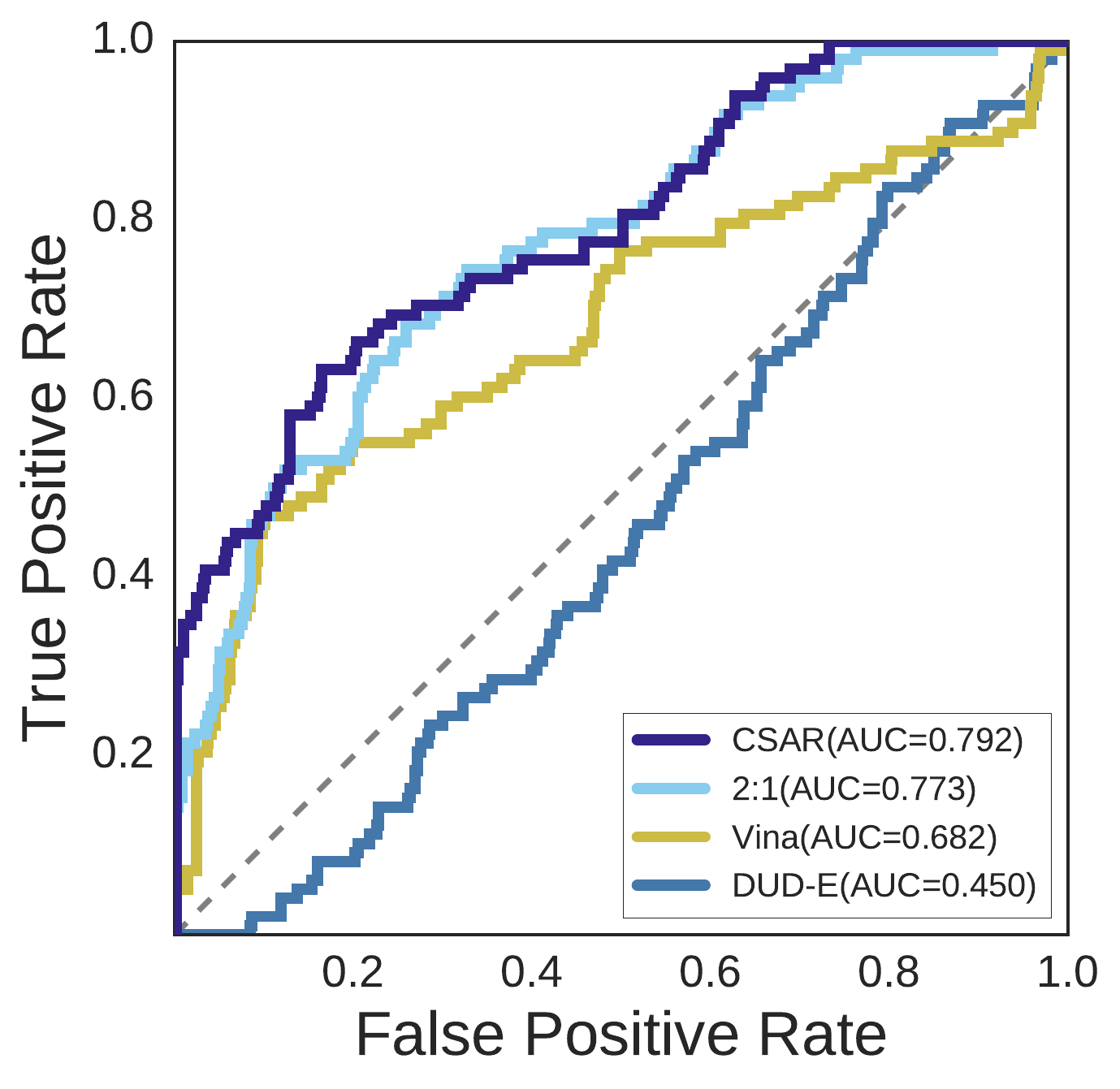}
\caption[]{\label{pdbbind_roc} ROC plot for discriminating low RMSD from high RMSD poses generated from the PDBbind core set. The CSAR-trained CNN performs best at classifying generated poses as low or high RMSD across targets, with a steep initial slope evincing good performance at early recognition.}
\end{figure}

\begin{figure}[tbp]
\includegraphics[width=0.5\linewidth]{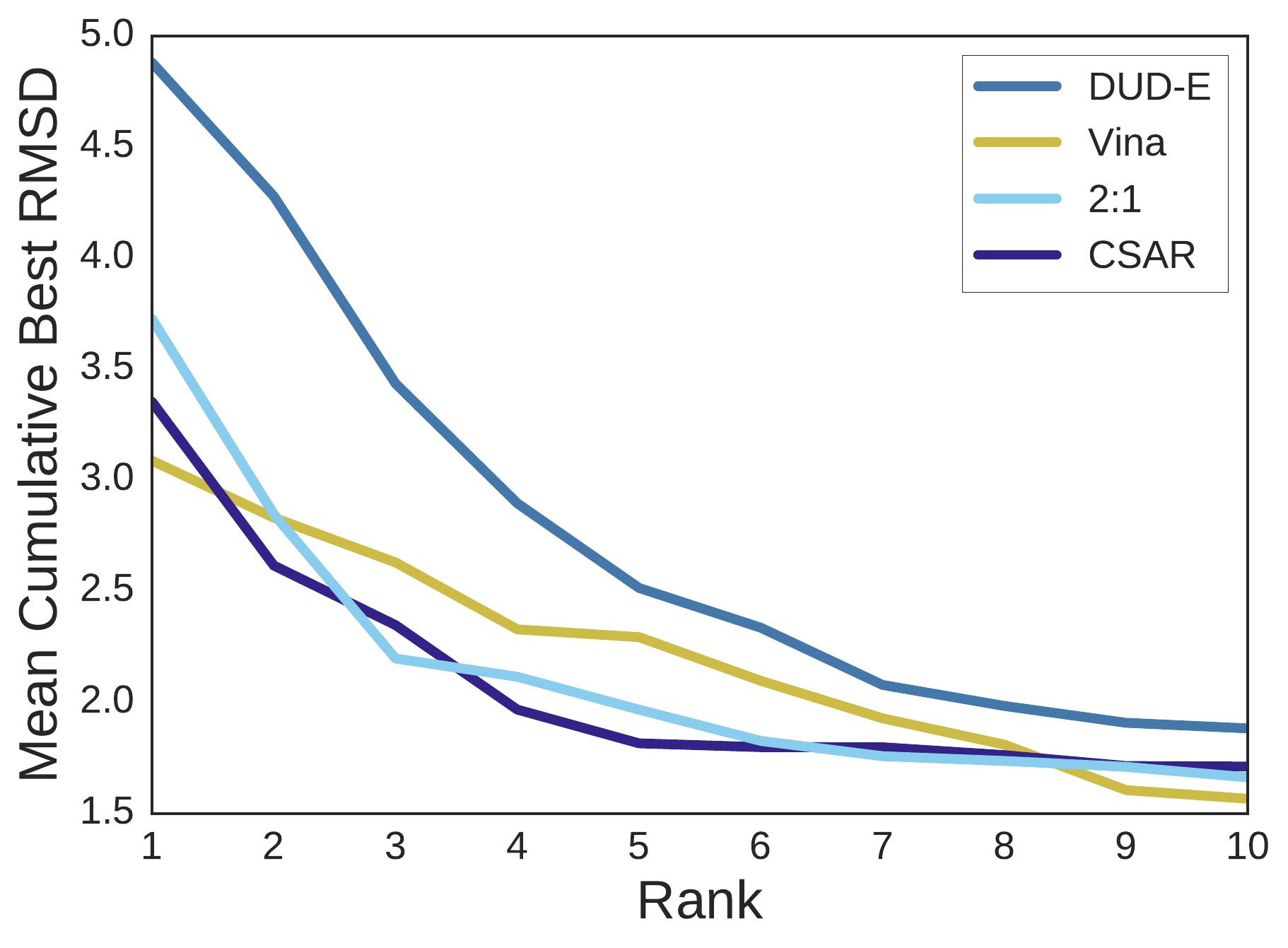}
\caption[]{\label{core_cumulbest} 
The best RMSD pose identified in the top ranked ligand poses averaged across all PDBbind core subset targets for each scoring method.
 }
\end{figure}

\begin{figure}[tbp]
\includegraphics[width=0.6\linewidth]{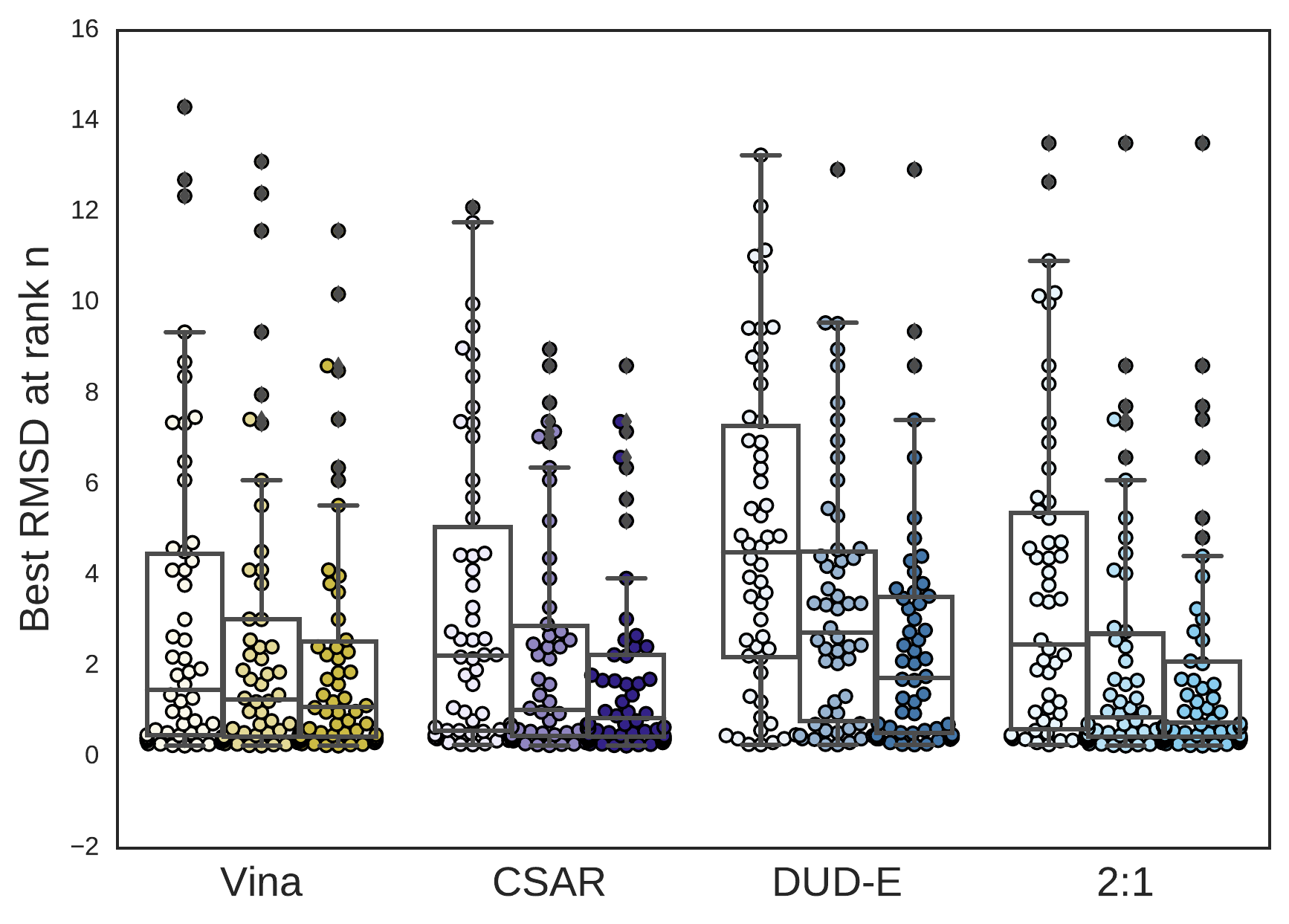}
\caption[]{\label{core_boxplot} Boxplots of the best RMSD seen so far at ranks 1, 3, and 5 (shown from left to right) for all targets in the PDBbind core subset.}
\end{figure}

\begin{figure}[tbp]
\includegraphics[width=0.6\linewidth]{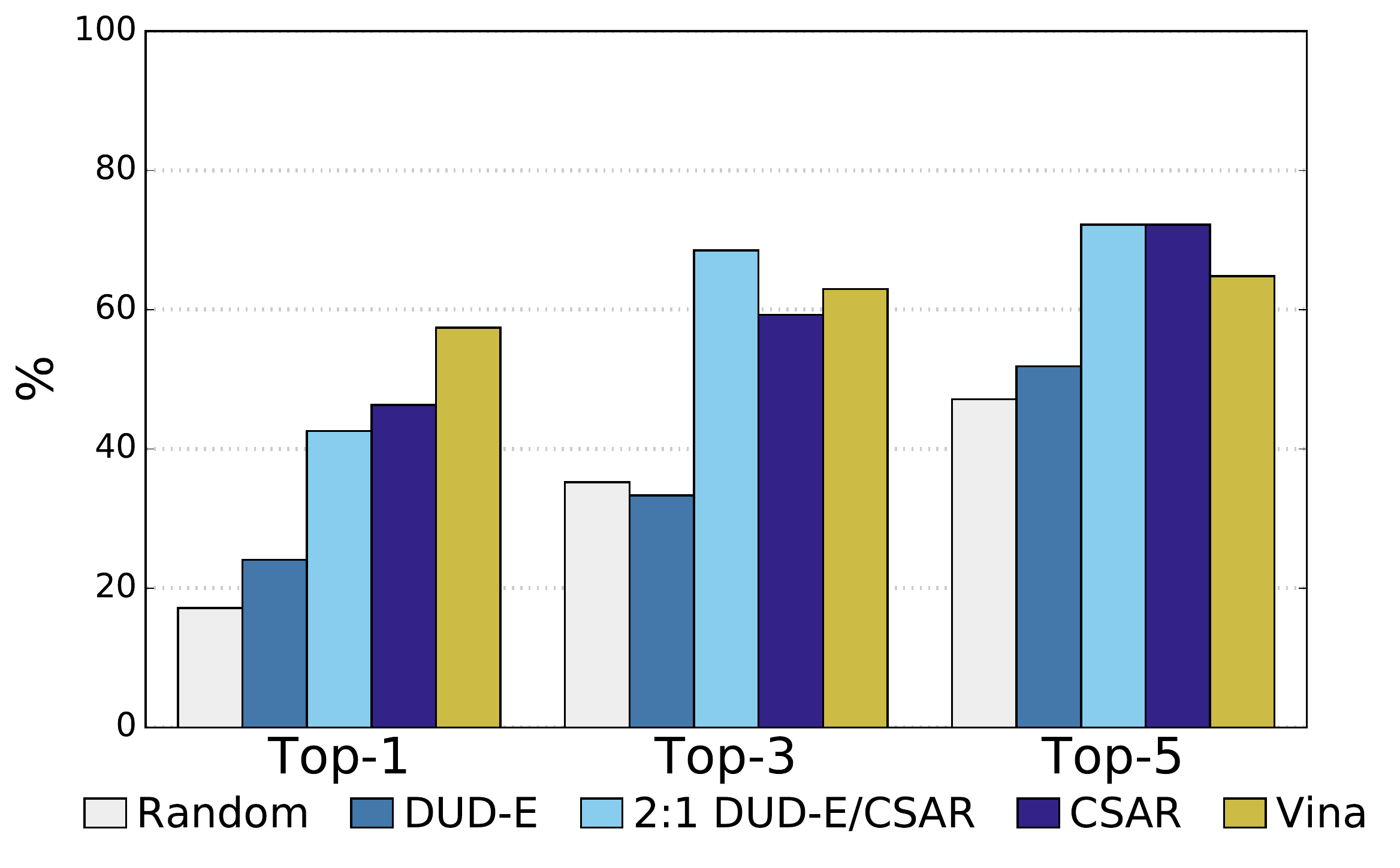}
\caption[]{\label{pdbbindpercents}The percentage of complexes with low RMSD poses identified as the top one, three or five poses for different scoring methods.}
\end{figure}

\begin{figure}[tbp]
\centering
\subcaptionbox*{3PE2}[0.32\linewidth]{
\includegraphics[width=0.32\linewidth]{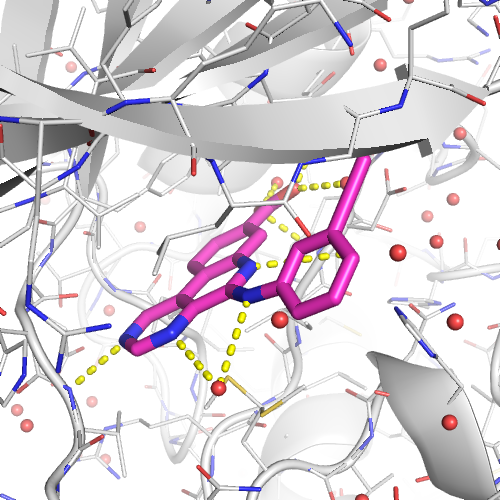}
}
\hfill
\subcaptionbox*{Vina}[0.32\linewidth]{
\includegraphics[width=0.32\linewidth]{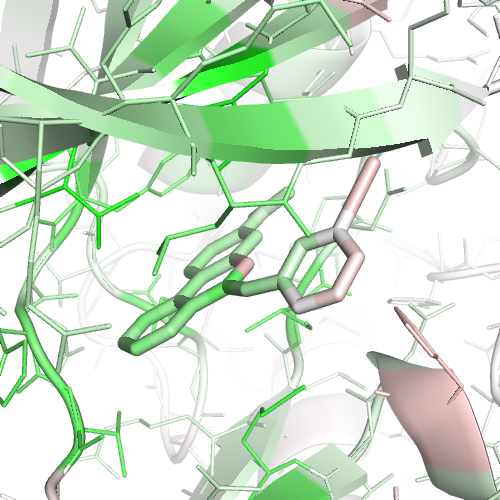}
}
\hfill
\subcaptionbox*{CNN}[0.32\linewidth]{
\includegraphics[width=0.32\linewidth]{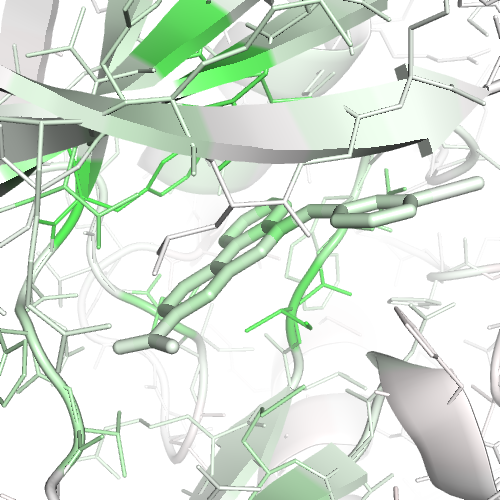}
}%
\caption{\label{vinagood} An example, PDB 3PE2, of a complex from the PDBbind core set where Vina correctly top-ranks a low RMSD pose  (0.25{\AA}) and the CNN model does not (5.27{\AA}).
The crystal pose is shown as magenta sticks and the two docked poses are visualized using the CSAR trained CNN model.}
\end{figure}

\begin{figure}[tbp]
\centering
\subcaptionbox*{3MYG}[0.32\linewidth]{
\includegraphics[width=0.32\linewidth]{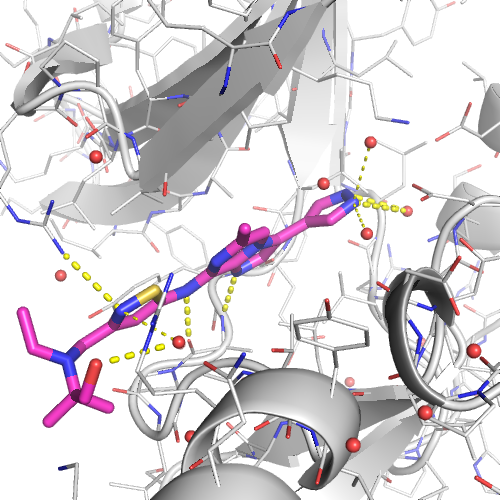}
}
\hfill
\subcaptionbox*{Vina}[0.32\linewidth]{
\includegraphics[width=0.32\linewidth]{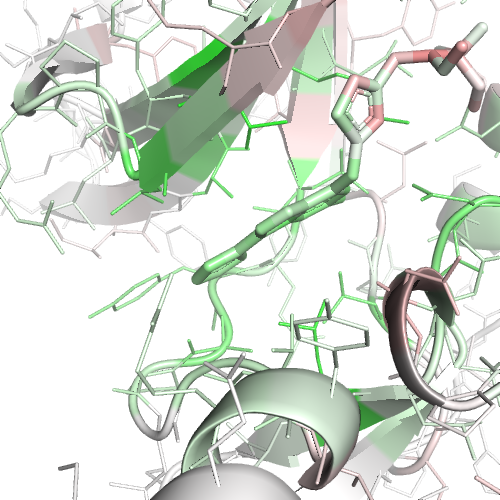}
}
\hfill
\subcaptionbox*{CNN}[0.32\linewidth]{
\includegraphics[width=0.32\linewidth]{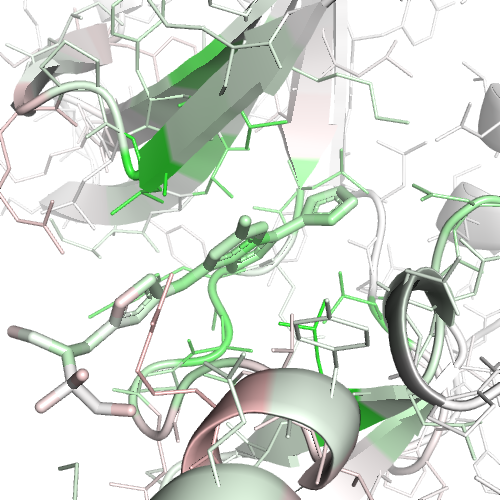}
}%
\caption{\label{vinabad} An example, PDB 3MYG, of a complex from the PDBbind core set where the CNN model correctly top-ranks a low RMSD pose (0.96{\AA}) and Vina does not (12.71{\AA}).
The crystal pose is shown as magenta sticks and the two docked poses are visualized using the CSAR trained CNN model.}
\end{figure}

\subsubsection{Pose Prediction}
Summary results for the PDBbind core set are shown in Figures~\ref{pdbbind_roc}, ~\ref{core_cumulbest}, \ref{core_boxplot}, and~\ref{pdbbindpercents}. As with the cross-validation results, the CNN models outperform Vina in an inter-target assessment of pose ranking (Figure~\ref{pdbbind_roc}) with an improvement of about 0.1 AUC. Also consistent with the cross-validation results is the finding that, on average, Vina's top-ranked pose has a lower RMSD than the top-ranked poses of any of the CNN methods, but by the second ranked pose the CSAR and DUD-E/CSAR combined CNN models, both of which were trained on pose prediction data, have improved on Vina (Figure~\ref{core_cumulbest}).  As expected, the model trained on DUD-E data, which consisted of inaccurate docked poses, does poorly at pose prediction.

The distribution of best RMSD values at different ranks is shown in Figure~\ref{core_boxplot}.  Even for the poorly performing DUD-E-only model there is a significant cluster of low RMSD poses. The percentage of complexes where a low RMSD pose ($<2${\AA}) was found in the top $N$ ranked poses for each method is shown in Figure~\ref{pdbbindpercents}. The DUD-E trained model had similar performance to random pose selection, providing further evidence for the conclusion that models trained on this kind of data lack pose sensitivity. The models trained with pose prediction data did significantly better than random, with the CSAR-trained model correctly identifying a low RMSD pose as the top ranked pose in 46\% of the complexes, compared to 57\% for Vina. As with the cross-validation results, accuracy improved significantly as the number of top ranked poses considered increased. The combined DUD-E/CSAR model outperformed Vina at identifying a low RMSD pose within the first three ranked poses. 

Examples of PDBbind poses visualized with the CSAR model are shown in Figures~\ref{vinagood} and~\ref{vinabad}.  Figure~\ref{vinagood} shows human protein kinase CK2 (PDB 3PE2).  For this complex, Vina correctly top-ranks a low RMSD pose while the CNN model prefers to flip the compound in the binding site.  The visualization illustrates why.  The CNN model correctly favors the binding of the low RMSD pose to the hinge region of the kinase (as indicated by the green highlighting on both the ligand and protein in this region), but it disfavors the position of the alkynyl.  Although flipping the compound results in less favorable interactions with the hinge region, it results in what the model considers to be a better pose of the alkynyl. Figure~\ref{vinabad} shows an Aurora A kinase (PDB 3MYG).  In this case, the CNN model correctly top-ranks a low RMSD pose while Vina prefers a pose that is flipped and more buried in the binding site.  Again, the model highlights the interactions with the hinge region of the kinase.  While the model slightly disfavors the solvent exposed portion of the compound, flipping the compound and burying this portion of the compound in the interior of the kinase is more strongly disfavored (as indicated by the red highlighting).

\begin{figure}[tbp]
\includegraphics[width=\linewidth]{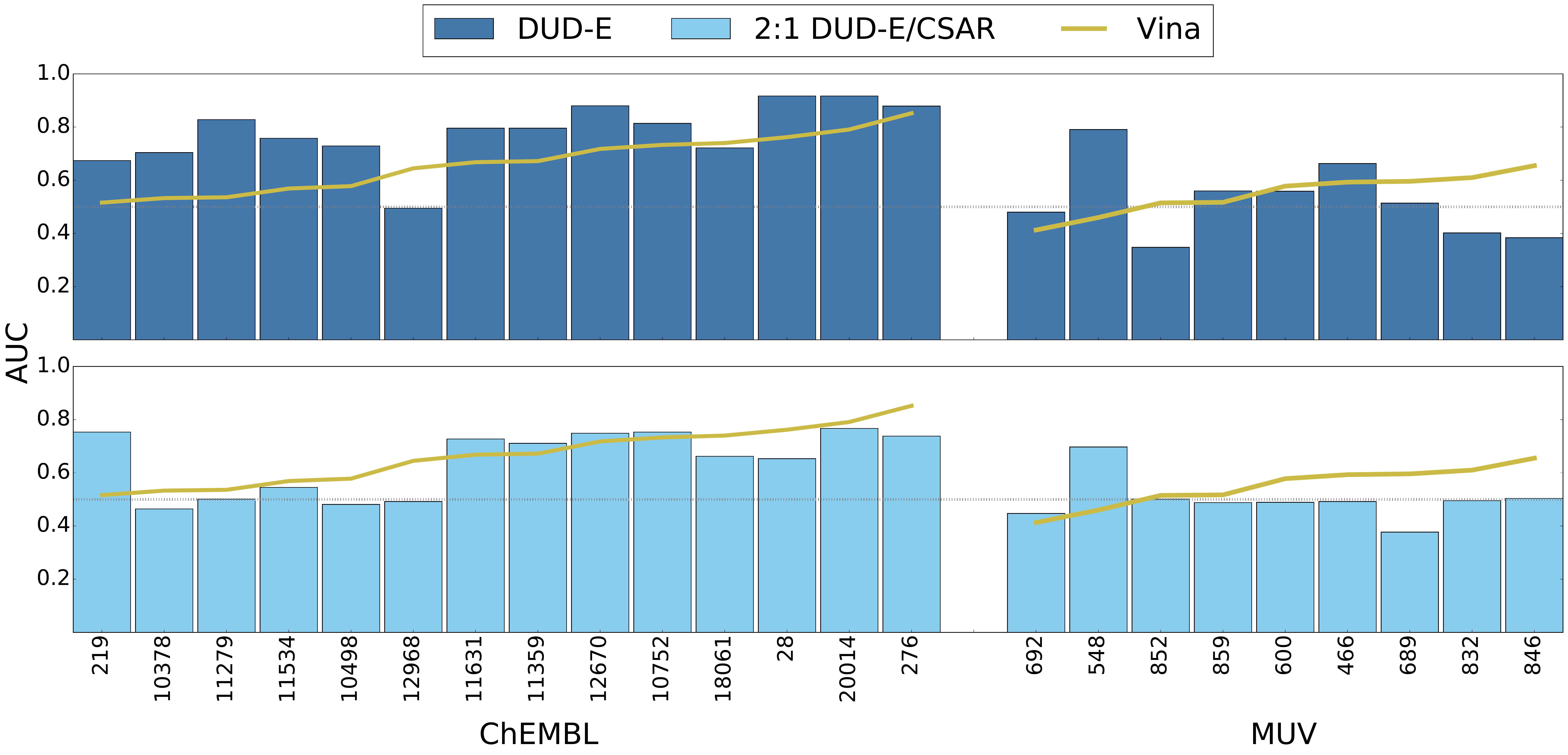}
\caption[]{\label{chemblmuvbars} Performance of CNN models on ChEMBL and MUV screening benchmarks
compared to the Vina scoring function. Targets are sorted by performance with Vina.
Identical sets of docked poses were ranked. The score of the top ranked pose of each ligand
is used to predict activity (multi-pose scoring). Consistent with the cross-validation results (Figure~\ref{bytarget}), a CNN model trained only on DUD-E training data performs best, outperforming Vina in 86\% of the ChEMBL targets and 56\% of the MUV targets. Models trained using a mix of DUD-E and CSAR data performed less well compared to Vina, achieving better AUCs than Vina in 36\% of the ChEMBL targets and 22\% of the MUV targets.
}
\end{figure}

\begin{figure}[tbp]
\centering
\subcaptionbox*{\label{chemblroc}ChEMBL}[0.48\linewidth]{
\includegraphics[width=0.48\linewidth]{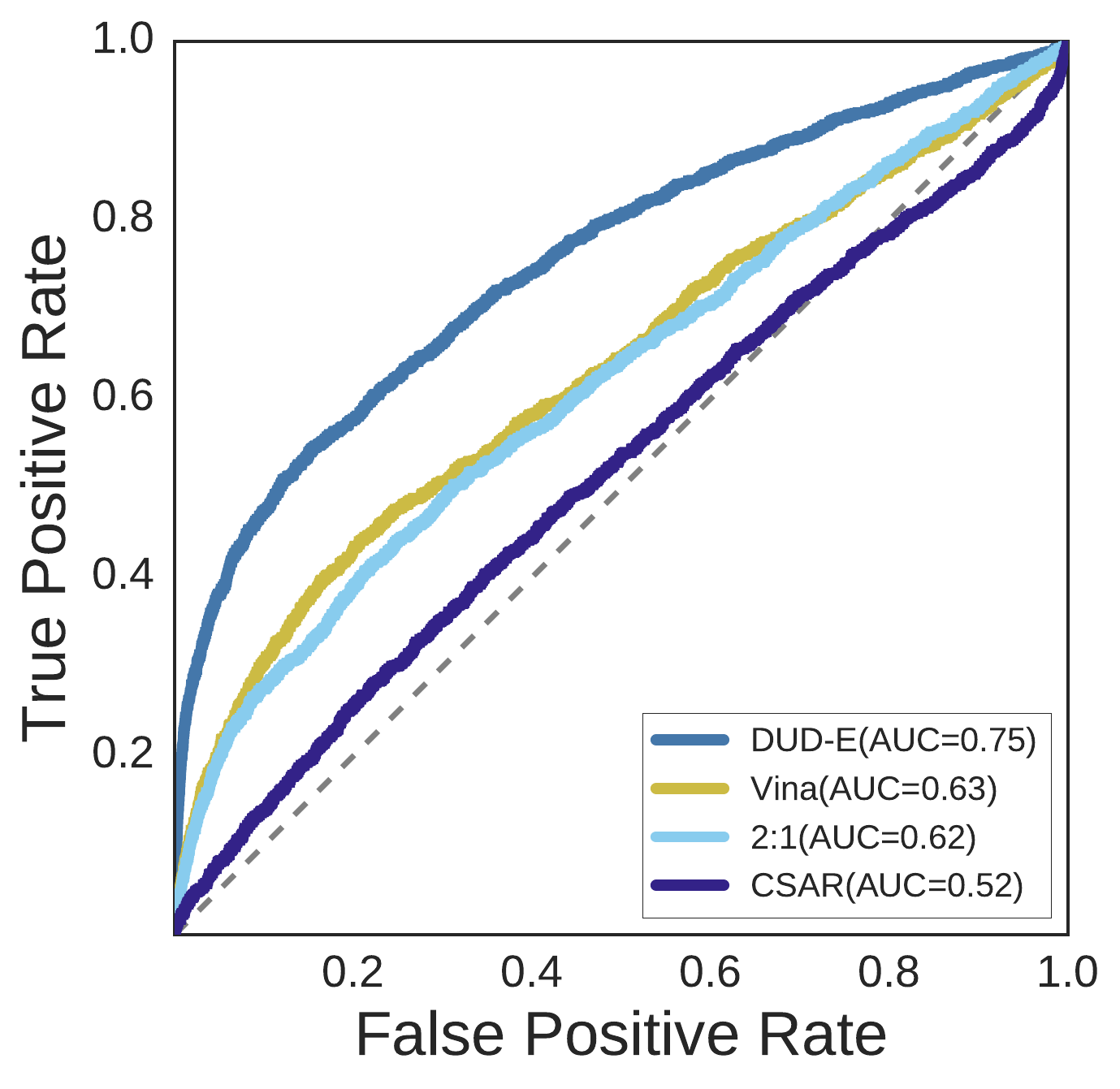}
}
\subcaptionbox*{\label{muvroc}MUV}[0.48\linewidth]{
\includegraphics[width=0.48\linewidth]{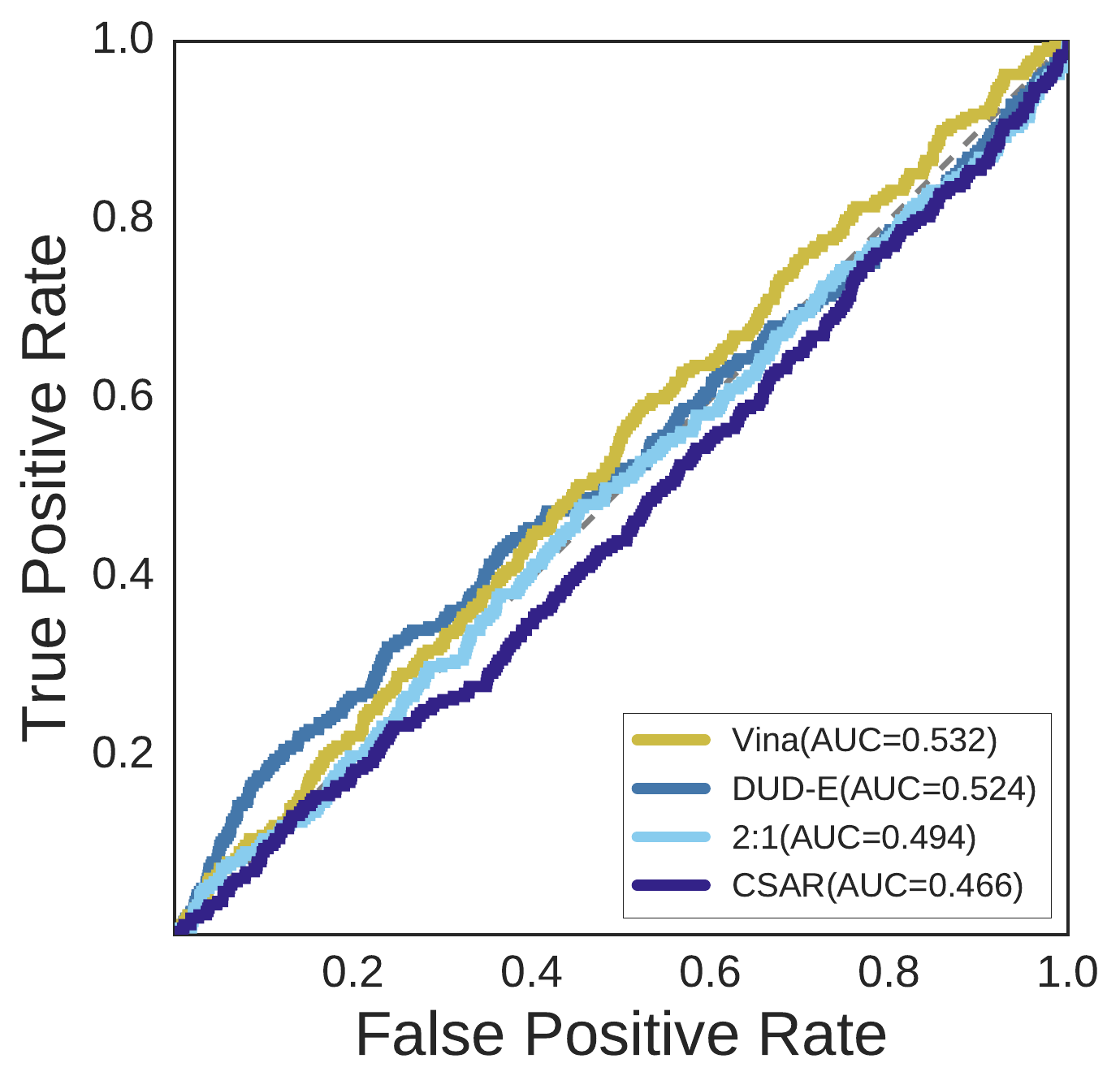}
}
\caption[]{\label{vsrocs} Overall virtual screening performance represented as a combined ROC curve for the three CNN models trained on the full training set and tested on the ChEMBL and MUV independent test sets and compared to Vina.}
\end{figure}

\begin{figure}[tbp]
\includegraphics[width=\linewidth]{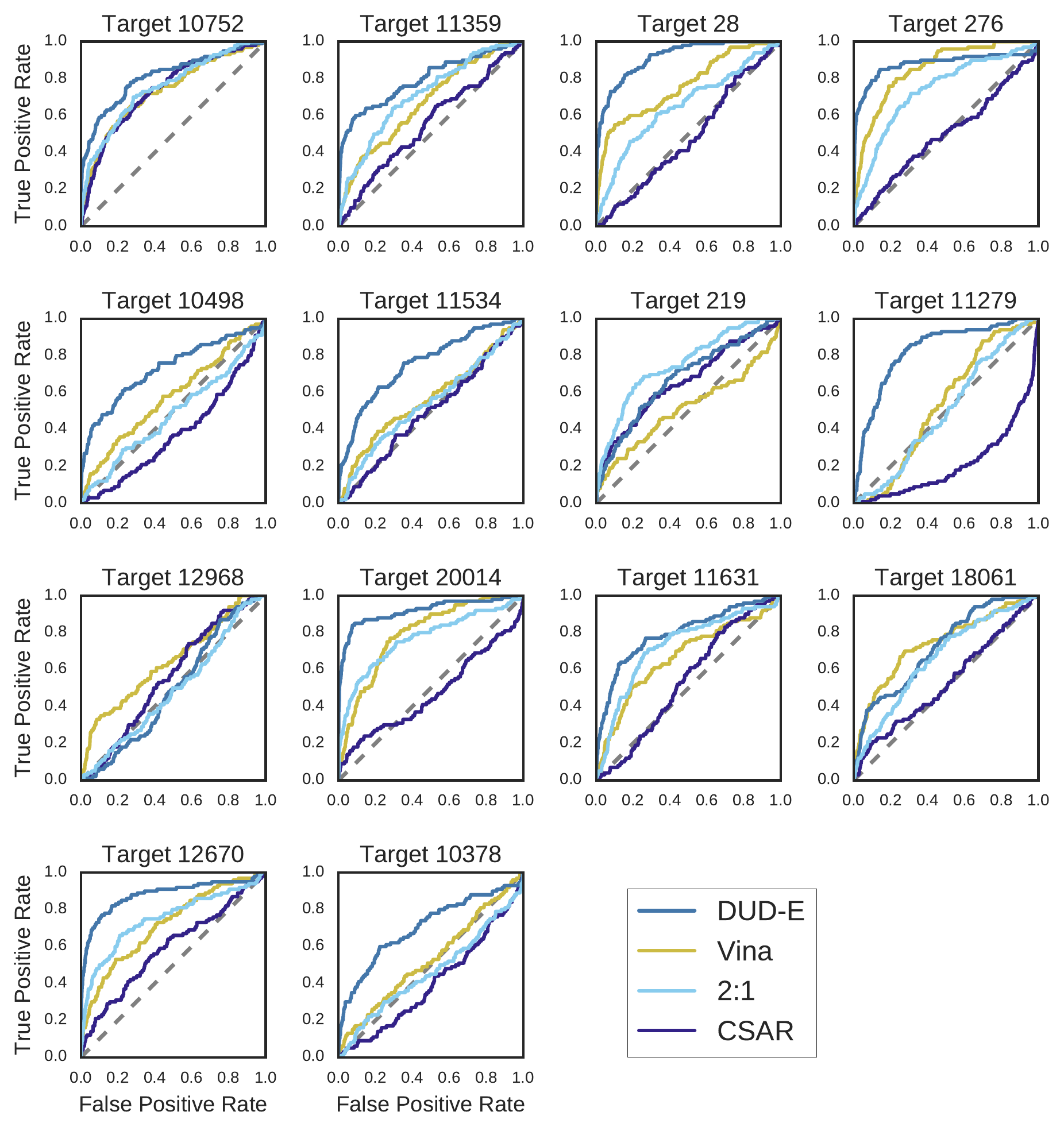}
\caption[]{\label{chembl_bytarget} Per-target ROC curves for Vina and the three CNN models for the ChEMBL test set.}
\end{figure}

\begin{figure}[tbp]
\includegraphics[width=\linewidth]{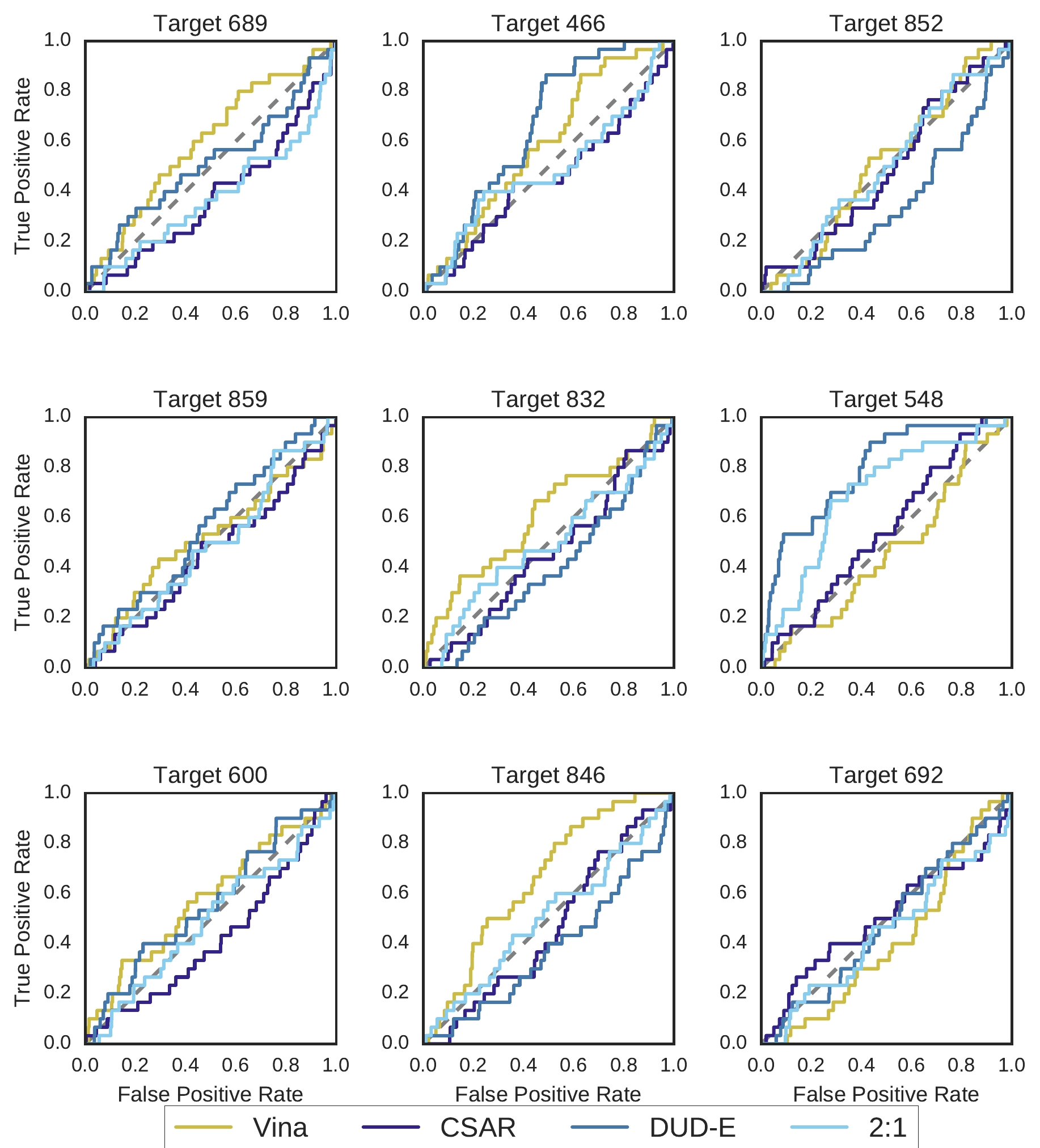}
\caption[]{\label{MUV_bytarget} Per-target ROC curves for Vina and the three CNN models for the MUV test set.}
\end{figure}

\begin{table}[tbp]
\begin{tabular}{r|c|c|c|c}
Target & Vina & DUD-E & 2:1 & CSAR \\ \hline \hline
219 & 0.516 & 0.674 & 0.753 & 0.657 \\
10378 & 0.533 & 0.704 & 0.464 & 0.402 \\
11279 & 0.536 & 0.828 & 0.501 & 0.213 \\
11534 & 0.569 & 0.758 & 0.545 & 0.497 \\
10498 & 0.578 & 0.729 & 0.481 & 0.380 \\
12968 & 0.645 & 0.495 & 0.492 & 0.561 \\
11631 & 0.668 & 0.796 & 0.727 & 0.54 \\
11359 & 0.672 & 0.796 & 0.711 & 0.554 \\
12670 & 0.718 & 0.880 & 0.749 & 0.590 \\
10752 & 0.733 & 0.814 & 0.753 & 0.739 \\
18061 & 0.740 & 0.722 & 0.662 & 0.531 \\
28 & 0.762 & 0.917 & 0.653 & 0.483 \\
20014 & 0.791 & 0.917 & 0.767 & 0.471 \\
276 & 0.852 & 0.879 & 0.738 & 0.496 \\
\end{tabular}
\caption{\label{chemblaucs} ChEMBL AUCs for Vina and CNN models trained on different training sets.}
\end{table}

\begin{table}[tbp]
\begin{tabular}{r|c|c|c|c}
Target & Vina & DUD-E & 2:1 DUD-E/ CSAR & CSAR \\ \hline \hline
692 & 0.413 & 0.480 & 0.447 & 0.505 \\
548 & 0.460 & 0.791 & 0.697 & 0.552 \\
852 & 0.515 & 0.348 & 0.501 & 0.491 \\
859 & 0.517 & 0.560 & 0.488 & 0.455 \\
600 & 0.578 & 0.559 & 0.489 & 0.422 \\
466 & 0.593 & 0.663 & 0.492 & 0.452 \\
689 & 0.596 & 0.514 & 0.377 & 0.381 \\
832 & 0.610 & 0.402 & 0.495 & 0.457 \\
846 & 0.655 & 0.384 & 0.504 & 0.461 \\
\end{tabular}
\caption{\label{muvaucs} MUV AUCs for Vina and CNN models trained on different training sets.}
\end{table}

\subsubsection{Virtual Screening}

Virtual screening results for the ChEMBL and MUV independent test sets are shown in Figures~\ref{chemblmuvbars}, \ref{vsrocs}, \ref{chembl_bytarget}, and~\ref{MUV_bytarget} and Tables~\ref{chemblaucs} and~\ref{muvaucs}. The ChEMBL and MUV tests sets are more challenging than the DUD-E benchmark for all methods.  The average AUCs for the ChEMBL benchmark are 0.67, 0.64, and 0.78 for the Vina, 2:1 DUD-E/CSAR CNN, and DUD-E CNN methods, which is consistently lower than the corresponding average cross-validation AUCs on DUD-E: 0.71, 0.80, and 0.86. Consistent with previously reported results\cite{riniker2013,Tiikkainen2009}, the MUV set is even more challenging with average AUCs of 0.55, 0.50, and 0.52 for Vina, 2:1 DUD-E/CSAR CNN, and DUD-E CNN.  Unlike the cross-validation results (Figure~\ref{mixed_data_bar}), the CSAR-trained CNN has close to random performance at virtual screening for most targets (Figures~\ref{vsrocs}, \ref{chembl_bytarget}, and~\ref{MUV_bytarget}).

Consistent with the cross-validation results, the DUD-E-trained CNN model generally outperforms the DUD-E/CSAR combined model. Since the ChEMBL and MUV sets were constructed using a methodology that differs from the DUD-E benchmark, this suggests that the DUD-E CNN model is learning genuinely useful information about features of the ligand and protein binding site that are relevant to binding, despite a lack of pose sensitivity, and is not learning an artificial artifact of the construction of the DUD-E set.  Interestingly, as shown in Figure~\ref{chemblmuvbars}, the targets with the biggest drop in performance between the DUD-E model and the pose sensitive DUD-E/CSAR model are also some of the targets with the lowest Vina performance. This would be the expected effect if docking is failing to sample accurate poses, as in this case a more cheminformatic-oriented, pose insensitive model would perform better.

\begin{table}[tbp]
\begin{tabular}{c|c|c|c|c}
Actives & Decoys & Vina & DUD-E & 2:1 \\ \hline \hline
MUV & MUV & 0.593 & 0.663 & 0.492 \\
MUV & ChEMBL & 0.619 & 0.682 & 0.523 \\
ChEMBL & ChEMBL & 0.668 & 0.796 & 0.727 \\
ChEMBL & MUV & 0.667 & 0.793 & 0.696\\
\end{tabular}
\caption{\label{decoyswap} The virtual screening performance for sphingosine 1-phosphate receptor EDG-1 (PDB 3V2Y) with different choices of active and decoy sets.  The active compounds were identified in different screens (biochemical for ChEMBL, cell-based for MUV) and the method used to construct the decoy sets is also different.}
\end{table}

The MUV benchmark is particularly challenging, with no method achieving an AUC greater than 0.6 on more than two targets. The overall performance across the benchmark is essentially random for all methods, as shown in Figure~\ref{vsrocs}.  Unlike with the ChEMBL set (Figure~\ref{chembl_bytarget}), in MUV the few individual targets where methods do appreciably better than random the improvement in AUC is not driven by early enrichment (Figure~\ref{MUV_bytarget}).  The use of cell-based assays and the lack of structures bound to ligands of known affinity (Table~\ref{testset_targets}) may make MUV a poor choice for a structure-based virtual screening assessment.  Alternatively, the observed poor performance may be due to the method MUV uses to construct the active and decoy sets, which attempts to avoid analog bias and artificial enrichment by ensuring that actives are well embedded in the chemical space of the decoys.  The MUV target 466, a lipid G protein-coupled receptor, is identical to ChEMBL target 11631, and we used the same structure, PDB 3V2Y, to generate poses.  This allows us to compare the effect of the different decoy construction approaches between the two benchmarks.  As shown in Table~\ref{decoyswap}, for all methods, the highest performance is achieved with the ChEMBL actives.  This suggests, for this target at least, that the method used to construct the decoys is not the cause of the observed poor performance and that the performance observed on the ChEMBL set is not due to artificial enrichment.

\begin{figure}[tbp]
	\includegraphics[width=450px]{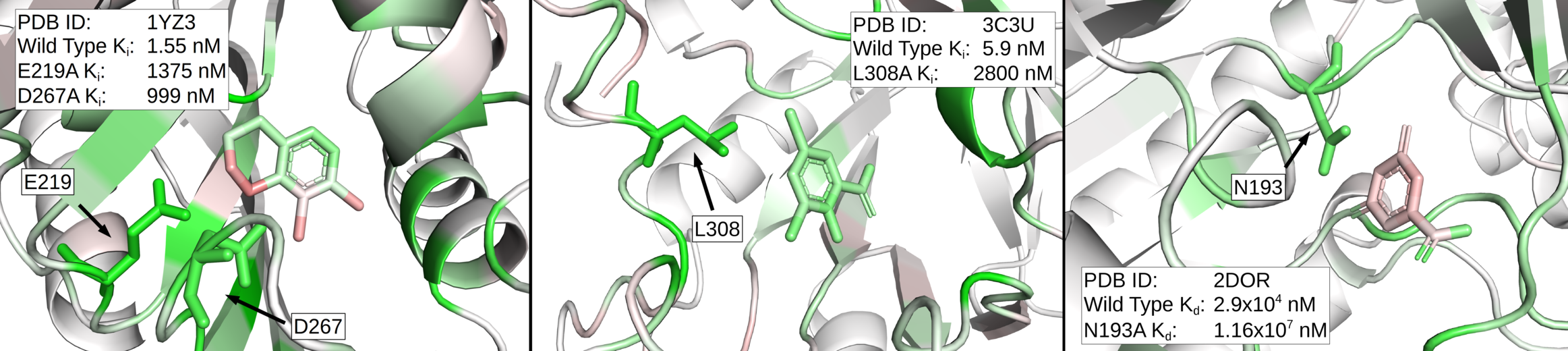}
    \caption{\label{mutfig}Visualizations of protein-ligand complexes with
    binding affinity data for point mutations in the protein. The top three most significant changes in binding
affinity from the Platinum database are shown from left to right. Any residue
that was mutated experimentally is shown in stick form, while the rest of the
protein is shown as a cartoon. In all three cases, the green coloring supports
the experimental results that the residues in question are important for ligand
binding.  Visualization is performed using the 2:1 DUD-E/CSAR model.}
\end{figure}

\subsection{Visualization}

Visualization is intended to provide a qualitative and easy to interpret indication of the atomic features that are driving the CNN model's output. In order to more quantitatively assess the utility of our visualization approach, we considered single-residue protein mutation data and partially aligned poses.

\subsubsection{Mutation Analysis}
The Platinum\cite{platinum} database provides
measured differences in protein-ligand binding affinity upon mutation of
single receptor residues. This experimental technique is a close analogue of the
visualization algorithm, where whole residues are removed and the complex re-scored.
For our assessment, we filtered the database to consider only experiments with single mutations to alanine or glycine in proteins that are not present in our training data and evaluated those with the largest changes in binding affinity.

The CNN was able to identify critical residues in many of the examples that
were tested. The three protein-ligand pairs with the highest changes in binding
affinity are shown in Figure~\ref{mutfig}. In all three cases, many residues had heavy green coloring, and the mutant residue is always colored green. Other highlighted residues may also be critical, but were not present in the Platinum database.  It is worth emphasizing that the CNN model was not trained on protein mutational data.  The fact that critical residues are highlighted suggests that the model is learning some general underlying model of the key features of protein-ligand interactions.

\begin{figure}[tbp]
\centering
\subcaptionbox*{3COY}[0.19\linewidth]{
\includegraphics[width=0.19\linewidth]{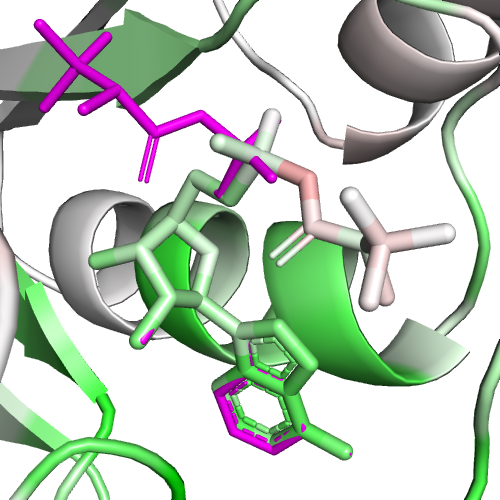}
}
\hfill
\subcaptionbox*{3UTU}[0.19\linewidth]{
\includegraphics[width=0.19\linewidth]{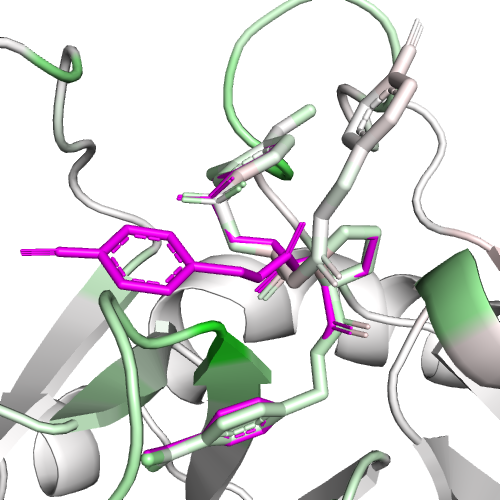}
}
\hfill
\subcaptionbox*{2QMJ}[0.19\linewidth]{
\includegraphics[width=0.19\linewidth]{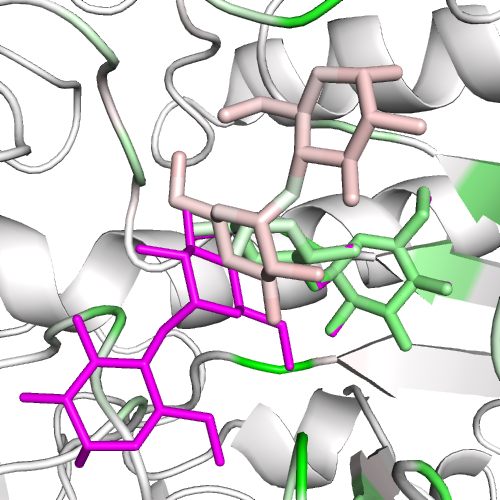}
}%
\hfill
\subcaptionbox*{3PWW}[0.19\linewidth]{
\includegraphics[width=0.19\linewidth]{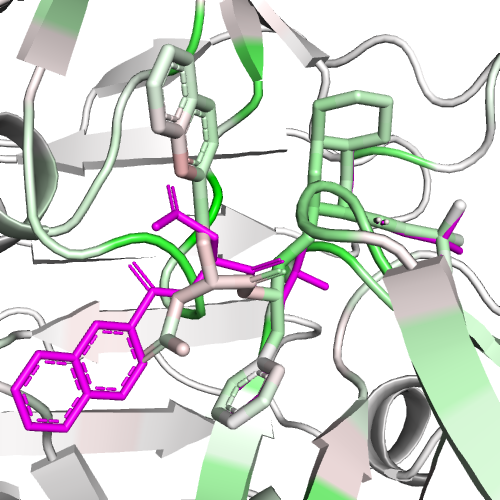}
}%
\hfill
\subcaptionbox*{3OZT}[0.19\linewidth]{
\includegraphics[width=0.19\linewidth]{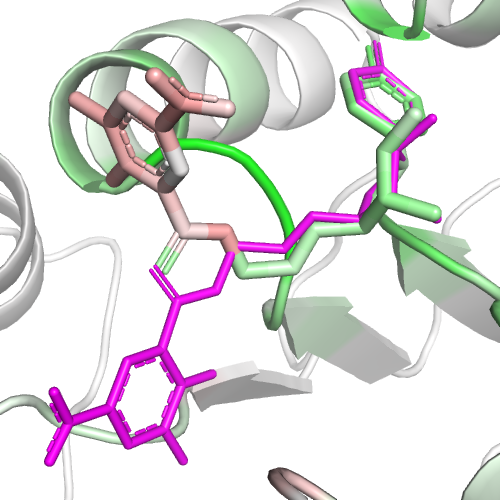}
}%
\caption{\label{partials} Visualizations of partially aligned docked poses from the PDBbind core set.  The crystal pose is shown as magenta sticks and the docked pose and receptor are colored according to our visualization algorithm and the 2:1 DUD-E/CSAR model.  None of these protein targets were included in training.  The visualization highlights that the model assesses the part of the pose aligned to the crystal ligand as more favorable than the differing part.}
\end{figure}

\subsubsection{Partially Aligned Poses}

We identified the high RMSD ($>4${\AA}) docked poses in the core PDBbind dataset that had the highest percentage of aligned atoms ($<0.1${\AA} distant to the corresponding crystal atoms).  These are poses that are partially correct; part of the molecule matches the crystal and part does not.

The five poses with the highest percentage of congruent atoms are shown visualized using the 2:1 DUD-E/CSAR model in Figure~\ref{partials}.  For all five poses, the CNN model ranks the crystal pose higher than the docked pose.  Our visualization shows why these poses are scored lower.
In all cases, the part of the docked pose that is aligned to the crystal pose is predominantly or entirely green (indicating positive contributions), but the divergent part of the ligand is entirely or partially red (indicating negative contributions).

\section{Discussion}

We have provided the first detailed description and evaluation of applying deep learning and convolutional neural networks to score protein-ligand interactions using a direct, comprehensive 3D depiction of the complex structure as input.  In many respects, our CNN models outperform standard approaches, exemplified here by the Autodock Vina scoring function.  In inter-target evaluations of pose prediction, both using cross-validation and an independent test set, CNN models can perform substantially better (e.g., Figures~\ref{depth3_all_roc} and \ref{pdbbind_roc}).  Likewise, CNN models can do well in virtual screening evaluations (e.g., Figures~\ref{bytarget} and~\ref{chemblmuvbars}).  However, our results also point to weaknesses in the current method and opportunities for improvement.

Although the CNN models performed well in an inter-target pose prediction evaluation, they performed worse at intra-target pose ranking (e.g., Figures~\ref{ranking_bar}, \ref{core_cumulbest}, and \ref{core_boxplot}), which is more relevant to molecular docking.  It is likely that intra-target ranking could be improved by changing the training protocol to more faithfully represent this task.  For example, currently ligands are treated identically regardless of their affinity, as long as they fall below a threshold (10$\upmu$M). It is conceivable that a high RMSD pose of a high affinity ligand should legitimately be scored better than a low RMSD pose of a low affinity ligand, a distinction the current training protocol cannot make.  Incorporating the binding affinity as a component of training, or performing relation classification \cite{santos2015classifying}, which assesses the ability of the network to \textit{rank} rather than score poses, may significantly improve intra-target performance of CNN models.

Our models perform well in a clustered cross-validation evaluation of virtual screening on the DUD-E benchmark. However, this benchmark may be susceptible to artificial enrichment \cite{ramsundar:2015}, resulting in overly optimistic predictions of virtual screening performance. We believe that our use of clustered cross-validation, which not only avoids training on ligands of the same target but also all similar targets, should mitigate some of the artificial enrichment issues inherent in DUD-E.  Furthermore, our independent test sets both used an entirely different method of dataset construction than the DUD-E set.

Ideally the CNN models learn a generalizable model of protein-ligand binding from the training data.  However, our models' ability to generalize beyond the task inherent in the training data, while present, is limited (e.g. Figure~\ref{mixed_data_bar}).  This is further highlighted by that fact that our CNN scores do not correlate ($|R|$ < 0.1) with binding affinity data when evaluating the CSAR crystal poses. In contrast, Vina exhibits a modest correlation ($R$ = 0.37) on the same benchmark. That is, training to classify poses and active/inactive compounds does not generalize to the regression problem of binding affinity prediction.  We expect that CNN models trained on binding affinity data would provide substantially improved results on this task.  Furthermore, our experience training combined pose prediction and virtual screening models indicates that multiple data types can be integrated to generate effective multi-task models. Unfortunately, we have not yet observed instances where including multi-task training data resulted in a synergistic effect, improving the performance of all tasks, although such an effect has been observed in other domains \cite{ramsundar:2015}.

In total, we believe that the current work demonstrates the potential of convolutional neural network models of protein-ligand binding to outperform current state-of-the-art methods.  There remain many possible avenues for improving CNN models, such as training with larger datasets spanning a range of objectives (e.g. pose ranking, affinity prediction, virtual screening, etc.) related to ligand binding.  In order to aid in the development of more robust and higher performance CNN models, all of our code and models are available under an open source license as part of our gnina molecular docking software at \url{https://github.com/gnina}.

\begin{acknowledgement}

We thank Justin Spiriti and Alec Helbling for their input on the manuscript.
This work is supported by R01GM108340 from the National Institute of General Medical Sciences. The TECBio REU@Pitt program is supported by the National Science Foundation under Grant DBI-1263020 and is co-funded by the Department of Defense in partnership with the NSF REU program.

\end{acknowledgement}



\bibliography{biblio}

\end{document}